\documentclass[preprint,1p]{elsarticle}

\usepackage{amsmath}
\usepackage{amssymb}
\usepackage{hyperref}
\usepackage{todonotes} 
\usepackage{soul}
\usepackage{algorithm}
\usepackage{algorithmic}
\usepackage{amsthm}
\usepackage{booktabs}
\usepackage{graphicx} 
\usepackage{siunitx}
\usepackage{booktabs}
\usepackage{makecell}
\usepackage{subcaption}
\usepackage{placeins}

\theoremstyle{plain}

\begin{document}
\journal{UNSURE}
\title{DEF: Diffusion-augmented Ensemble Forecasting}

\author[1]{David Millard\corref{cor1}}
\ead{djm3622@rit.edu}

\author[2]{Arielle Carr}
\ead{arg318@lehigh.edu}

\author[3]{Stéphane Gaudreault}
\ead{stephane.gaudreault@ec.gc.ca}

\author[1]{Ali Baheri}
\ead{akbeme@rit.edu}

\cortext[cor1]{Corresponding author}

\address[1]{Kate Gleason College of Engineering, Rochester Institute of Technology, Rochester, USA}
\address[2]{Computer Science \& Engineering, Lehigh University, Bethlehem, USA}
\address[3]{Recherche en prévision numérique atmosphérique, Environnement et Changement climatique Canada, Dorval, Canada}




\begin{abstract}
\sloppy
We present DEF (\textbf{\ul{D}}iffusion-augmented \textbf{\ul{E}}nsemble \textbf{\ul{F}}orecasting), a novel approach for generating initial condition perturbations. Modern approaches to initial condition perturbations are primarily designed for numerical weather prediction (NWP) solvers, limiting their applicability in the rapidly growing field of machine learning for weather prediction. Consequently, stochastic models in this domain are often developed on a case-by-case basis. We demonstrate that a simple conditional diffusion model can (1) generate meaningful structured perturbations, (2) be applied iteratively, and (3) utilize a guidance term to intuitivey control the level of perturbation. This method enables the transformation of any deterministic neural forecasting system into a stochastic one. With our stochastic extended systems, we show that the model accumulates less error over long-term forecasts while producing meaningful forecast distributions. We validate our approach on the 5.625$^\circ$ ERA5 reanalysis dataset, which comprises atmospheric and surface variables over a discretized global grid, spanning from the 1960s to the present. On this dataset, our method demonstrates improved predictive performance along with reasonable spread estimates.
\end{abstract}

\maketitle


\section{Introduction}\label{sec:introduction}
In 2023, a machine learning model surpassed the best operational numerical weather prediction models. Pangu-Weather \cite{bi2022panguweather3dhighresolutionmodel}, a 3D Earth-Specific Transformer, demonstrated state-of-the-art accuracy in the medium-range forecast (10–15 days). More remarkably, it achieved this in a fraction of the computational cost\textendash producing forecasts in minutes on a GPU, whereas traditional models require custom hardware processing petaflops of operations over the course of hours. Since then, the machine learning weather prediction community has expanded, leading to the development of more advanced models such as GraphCast \cite{lam2023graphcastlearningskillfulmediumrange}, FourCastNet \cite{pathak2022fourcastnetglobaldatadrivenhighresolution}, and FuXi \cite{zhong2024fuxi20advancingmachinelearning}. Despite their architectural differences, they all share one trait: deterministic forecasting.

Two common approaches for introducing stochasticity to the trajectories of a forecast are ensemble methods and condition perturbation. Ensemble methods involve using multiple models with the same input to generate different trajectories, achieved by varying model parameters. While powerful, these methods are extremely expensive, particularly in machine learning, where training a model can take weeks to months. Condition perturbation, on the other hand, involves perturbing known conditions to generate plausible alternative states that the system could have experienced. These perturbations are both effective and cost-efficient, as they account for the inherent uncertainty in observations. However, the primary challenge lies in designing an effective perturbation strategy. Many modern condition perturbation techniques employ data assimilation methods, which leverage known observations and error estimates to generate conditions that theoretically maximize exploration of the forecasting space. While theoretically compelling, these methods require both a significant history of observational data and a large number of ensemble members to fully explore the space.

The machine learning weather prediction (MLWP) community has addressed the challenge of ensemble forecasting on a case-by-case basis. Following the success of GraphCast, GenCast \cite{price2024gencastdiffusionbasedensembleforecasting} was introduced\textendash the same GraphCast model trained as a conditional diffusion model. FourCastNet demonstrated a stochastic approach by employing an ensemble of deterministic forecasts. Other techniques, such as Monte Carlo dropout, conformal prediction, and Bayesian neural networks, offer a more general framework for uncertainty quantification within deep learning but face several practical limitations when applied to the sensitivity and complexities of modern MLWP architectures. This discrepancy arises because, unlike many other fields, neural forecasting systems are trained to be highly sensitive to their inputs. 

In this study, we present an approach that successfully incorporates all three aspects: low complexity, fast speed, and high transferability. Our method is modular in design, leverages the well-studied and widely understood conditional diffusion methodology, and, after applying various speedups from modern diffusion literature, can generate hundreds of perturbations within minutes. Utilizing this technique, any deterministic model can effectively generate a large ensemble of members without requiring modifications to inference or training, aside from simply perturbing the initial conditions and autoregressing upon them.

\noindent \textbf{Our Contributions.} The contributions of this paper are threefold:
\begin{itemize}
    \item We develop a novel method for condition perturbation by performing conditional generation on the input conditions themselves. This method requires significantly less computational power, as the model generates within its own state rather than predicting the next state.
    \item We demonstrate that the method can be applied iteratively to perform "random walks" within a condition, offering more exploration power and better distribution estimation of the extremely large trajectory space.
    \item We show that the guidance term, the scalar parameter $\omega$ used within classifier free guidance, intuitively controls the strength of the perturbation, directly regulating the level of uncertainty quantification and enabling the controlled exploration of the state space by informed users.
\end{itemize}

\noindent \textbf{Paper Organization.} The remainder of this paper is structured as follows. Section 2 is a literature review covering recent techniques used for uncertainty quantification. Section 3 establishes foundational concepts in ensemble forecasting, neural forecasting systems, and denoising diffusion probabilistic models. Section 4 presents the methodology, detailing our dataset, models, and training/inference procedures. Section 5 presents our numerical experiments/results, Section 6 discusses key findings, and Section 7 outlines future work and directions enabled by this study.

\section{Related Work}

Model ensembling is the most popular method for ensemble forecasting. Within the context of deep learning, the method relies on training multiple deterministic members and using their distribution of predictions. Due to the randomness of training deep learning methods, differing random seeds produce slightly different predictive models, and due to the sensitivity of forecasting models, they can lead to diverging results in the long term. FourCastNet \cite{pathak2022fourcastnetglobaldatadrivenhighresolution} proposed the use of 1,000 deterministic members to generate an effective ensemble that was able to compete with the ECMWF \cite{81149}, which has 30-50 members due to the significant computational demand of numerical weather prediction (NWP) \cite{BAER200091} schemes. Modern perspectives are now considering huge ensembles of 1,000 to 10,000 members \cite{mahesh2025hugeensemblesidesign} utilizing spherical Fourier neural operators \cite{bonev2023sphericalfourierneuraloperators}, an extension of FourCastNet that operates on the spherical coordinates of Earth rather than grid interpolations. While these approaches are very promising, they all suffer from a lack of probabilistic formulation and typically require a significant number of models to ensure the empirical distribution is large enough to effectively capture the uncertainty of the dynamics, resulting in an ensemble underdispersion.

One established machine learning architecture for generating probabilistic forecasts is Bayesian neural networks \cite{charnock2020bayesianneuralnetworks}. These networks modify the conventional neural network framework by learning distributions of weights rather than point estimates. This modification fundamentally alters the architecture and training procedures but demonstrates significant advantages when modeling inherently uncertain data. Recent research has specifically examined the application of this approach for predicting North Atlantic sea surface temperature using a convolutional LSTM architecture \cite{https://doi.org/10.1029/2022MS003058}. The investigators demonstrated that through their probabilistic formulation, the direct point estimate of the prediction exhibited superior performance on average compared to an equivalent model with standard deep learning weight parameterization. Furthermore, the study presents a quantitative comparison between a conventional deep learning ensemble and their SVGD ensemble derived from the Bayesian convolutional LSTM. Their analysis reveals that the SVGD methodology mitigates the ensemble collapse phenomenon frequently observed in standard deep learning ensembles, wherein ensemble members produce nearly identical predictions with minimal variance. Although Bayesian neural networks represent a promising approach for ensemble forecasting, they face significant scalability limitations. Despite advances in addressing these constraints \cite{pmlr-v108-farquhar20a}, current methodologies remain insufficient to scale to the computational requirements necessary for global neural ensemble forecasting applications.

 Initial condition perturbation is one of the most extensively studied methods in ensemble forecasting within the weather prediction community. The simplest approach is noise addition, which lacks a physical interpretation and is instead based solely on the uncertainty of signal readings. Another method, time lagging \cite{ShortRangeNumericalWeatherPredictionUsingTimeLaggedEnsembles}, leverages a model's past forecasts as conditions for the current system. As a model produces multistep forecasts further into the future, it explores one trajectory that deviates from the true trajectory. Once ground truth conditions are observed, previous forecasts can be incorporated into the ensemble. This method is computationally efficient, as it reuses prior forecasts without additional computation. However, it performs poorly \cite{DifferentInitialConditionPerturbationMethodsforConvectionPermittingEnsembleForecastingoverSouthChinaduringtheRainySeason} since it relies on point estimates that accumulate error over time. Additionally, time lagging is highly dependent on the model itself\textendash when the model underperforms, this method does not aid in learning a true distribution. Bred vectors \cite{ComparisonbetweenSingularVectorsandBreedingVectorsasInitialPerturbationsfortheECMWFEnsemblePredictionSystem}, a data assimilation technique, offer a dynamic alternative by leveraging the natural instabilities of the system. The breeding method perturbs an initial state, integrates it forward using the full nonlinear model, and periodically rescales the perturbations to maintain a fixed amplitude. This iterative process ensures that the perturbations align with the dominant growing modes of the system. A related technique, singular vectors \cite{ComparisonbetweenSingularVectorsandBreedingVectorsasInitialPerturbationsfortheECMWFEnsemblePredictionSystem}, perturbs the singular values of the system to ensure that perturbations occur along the dominant modes. It achieves this by computing the singular value decomposition of the tangent linear model of the full nonlinear system, perturbing the singular values, and transforming back. While data assimilation methods such as these represent the state of the art, they suffer from two distinct limitations: they require a long history of predictions, observations, and errors, and they are designed for NWP models. Within the MLWP setting, we lack a direct method for time integration or obtaining tangent linear models. As a result, applying these techniques would require substantial reformulation and generous assumptions.

Given the explosion of generative AI and its long track record of success, variational autoencoders (VAEs) \cite{Kingma_2019} and denoising diffusion probabilistic models (DDPMs) \cite{ho2020denoisingdiffusionprobabilisticmodels} have increasingly been applied to tasks within the MLWP community, such as extreme ERA5 compression \cite{han2024cra5extremecompressionera5} and upscaling \cite{merizzi2024windspeedsuperresolutionvalidation}. A particularly notable and influential work in this direction is GenCast \cite{price2024gencastdiffusionbasedensembleforecasting}, which serves as a significant inspiration for this study. GenCast is a conditional diffusion model trained for stochastic next-state prediction. Architecturally, it is identical to GraphCast \cite{lam2023graphcastlearningskillfulmediumrange} but differs in its training objective. Specifically, GenCast conditions the graph-based model on the previous state, formally defined as $G_\theta(x_t \mid x_{t-1})$. Given its formulation under DDPMs, the authors have found GenCast to be among the most accurate forecasting models, surpassing even ECMWF ensembles and AFNO-based large ensembles. Notably, because GenCast modifies the training objective rather than the architecture, it allows for seamless integration into existing forecasting pipelines, making it highly adaptable for further experimentation.

\section{Background}

\subsection{Ensemble Forecasts}
Given an model $\mathcal{M}$, an ensemble forecast consists of a set of predictions $\{X_i(T)\}_{i=1}^{N}$, where each $X_i(T)$ represents the forecasted state at time $T$ from an initial condition $X_i(0)$. The ensemble mean, which provides an averaged forecast, is given by:
\begin{equation}
\bar{X}(T) = \frac{1}{N} \sum_{i=1}^{N} X_i(T),
\end{equation}
where $T$ is the current timestep and $N$ is the number of members. The ensemble spread, representing forecast uncertainty, is defined as:
\begin{equation}
S(T) = \sqrt{\frac{1}{N} \sum_{i=1}^{N} \left( X_i(T) - \bar{X}(T) \right)^2 }.
\end{equation}
The spread-skill relationship suggests that ensemble spread $S(T)$ should correlate with forecast error such that $S(T) \propto \|\bar{X}(T) - X(T)\|_2$, providing a measure of reliability in probabilistic forecasts.

Given the relationship $S(T) \propto \|\bar{X}(T) - X(T)\|_2$, we observe an inverse problem where maximizing the forecast spread leads to a larger forecast error. As a result, it is common practice to aim for $S(T) \approx \|\bar{X}(T) - X(T)\|_2$, providing a clearer goal. This observation can be adjusted to accommodate specific needs, but we consider it an ideal benchmark in the results section of our work. This metric is prominently reflected in two key evaluation criteria: the Continuous Ranked Probability Score (CRPS) \cite{lang2024aifscrpsensembleforecastingusing} and energy scoring \cite{zheng2025mvgcrpsrobustlossfunction}, both of which directly quantify forecast spread and skill. Given these methods for evaluating ensemble forecast performance, there exist two primary approaches for constructing ensembles: condition-based and model-based methods.

\subsection{Condition Perturbations}
Condition perturbations are a natural method for forecasting. This type of ensemble leverages the uncertainty in observations. Although we typically refer to our observations as "ground truth," the process of collecting these observations is imperfect and can introduce instabilities. Given the sensitivity of a model $\mathcal{M}$, we assume an observation $X(T)$ is drawn from a distribution of possible observations:
\begin{equation}
X(T) \sim \mathcal{P}(X | \mathcal{M}, \mathcal{O}),
\end{equation}
where $\mathcal{P}(X | \mathcal{M}, \mathcal{O})$ represents the conditional distribution of observations given the model $\mathcal{M}$ and the underlying observation process $\mathcal{O}$, which accounts for measurement noise and uncertainty. We can then introduce perturbations as follows:  
\begin{equation}
X_i(T) \sim F(X(T)), \quad i = 1, \dots, N,
\end{equation}
where \( F(X(T)) \) represents a perturbation function that generates ensemble members by introducing controlled variations to the initial conditions. These perturbed states \( \{X_i(T)\}_{i=1}^{N} \) serve as the initial conditions for several ensemble members of a forecasting model \( G_\theta \). Given that our model \( G_\theta \) approximates the underlying dynamics, we assume  
\begin{equation}
X(T+1) \approx G_\theta(X(T)),
\end{equation}  
where \( G_\theta \) is expected to produce stable forecasts under near-zero perturbations and $F$ is the actual dynamics. However, if the initial conditions exhibit small but non-negligible differences, i.e.,  
\begin{equation}
X_i(T) \not\approx X_j(T), \quad \forall i \neq j,
\end{equation}  
these perturbations can amplify over time, leading to significant divergence in forecast trajectories across ensemble members. This divergence enables a robust estimation of both the ensemble mean \( \bar{X}(T) \) and the forecast spread \( S(T) \), effectively quantifying the inherent uncertainty in the system.  

\subsection{Neural-based Forecasting}
Neural-based forecasting methods utilize large deep learning models trained on historical data to predict atmospheric states. Unlike conventional Numerical Weather Prediction (NWP) models, which numerically solve partial differential equations (PDEs), neural models learn a mapping between the current state and the forecasted state at the next time step:
\begin{equation}
G_{\phi}: X_t \rightarrow X_{t+\Delta t},
\end{equation}
where \( G_{\theta} \) is a neural network parameterized by \( \theta \), and \( X_t \) represents the atmospheric state at time \( t \). These models typically use a regressive loss function as their objective, such as mean absolute error (MAE), mean squared error (MSE), or root mean squared error (RMSE). However, they can also incorporate physical constraints via Physics-Informed Neural Networks (PINNs) \cite{raissi2024physicsinformedneuralnetworksextensions}. In this approach, the loss function is enhanced with a physics-based term that enforces consistency with known governing equations:
\begin{equation}
\mathcal{L} = \mathcal{L}_{\text{data}} + \lambda \mathcal{L}_{\text{physics}},
\end{equation}
where \( \mathcal{L}_{\text{data}} \) is the standard regression loss, \( \mathcal{L}_{\text{physics}} \) penalizes violations of the governing equations, and \( \lambda \) is a hyperparameter that balances the two objectives.

Currently, state-of-the-art neural forecasting systems include GraphCast, FourCastNet, and Pangu-Weather. GraphCast is a graph neural network (GNN)-based model \cite{zhou2021graphneuralnetworksreview} that represents weather fields as a graph \( \mathcal{G} = (V, E) \), where \( V \) are spatial nodes and \( E \) are edges encoding dependencies between nodes. Forecasts are computed through message passing:
\begin{equation}
h_v^{(l+1)} = \sigma \left( W^{(l)} h_v^{(l)} + \sum_{u \in \mathcal{N}(v)} W^{(l)} h_u^{(l)} \right),
\end{equation}
where \( h_v^{(l)} \) is the hidden representation of node \( v \) at layer \( l \), and \( \mathcal{N}(v) \) denotes the neighbors of node \( v \). FourCastNet employs Adaptive Fourier Neural Operators (AFNOs) \cite{guibas2022adaptivefourierneuraloperators}, which perform global convolutions in the spectral domain:
\begin{equation}
h_{i} = \mathcal{F}^{-1} \left( \sigma \left( W \mathcal{F}(X_t) \right) \right),
\end{equation}
where \( \mathcal{F} \) is the Fourier transform and \( W \) is a learned weight matrix. Pangu-Weather utilizes a Transformer-based \cite{vaswani2023attentionneed} architecture that models atmospheric dynamics via self-attention:
\begin{equation}\label{eq:attn}
Z = \text{softmax} \left( \frac{QK^T}{\sqrt{d}} \right) B,
\end{equation}
where \( Q, K, B \) are the query, key, and value matrices, respectively, and \( d \) is the feature dimension. In this paper, we explore a simpler yet powerful approach using an attention U-Net model \cite{oktay2018attentionunetlearninglook}. Attention U-Net augments the standard U-Net architecture \cite{ronneberger2015unetconvolutionalnetworksbiomedical} by incorporating an attention mechanism, as described in Equation \ref{eq:attn}, over the entire spatial domain. To further enhance efficiency within our computational constraints, we limit the attention mechanism to the lower two downsampled spaces. This allows us to maintain a balance between performance and computational cost while preserving the benefits of the attention mechanism.

\subsection{Denoising Diffusion Probabilistic Models}
Denoising Diffusion Probabilistic Models (DDPMs) represent a class of generative models that learn to reverse a gradual noising process. The forward diffusion process transforms data samples into pure noise through a fixed Markov chain of $T$ steps:
\begin{equation}
q(x_t | x_{t-1}) = \mathcal{N}(x_t; \sqrt{1 - \beta_t} x_{t-1}, \beta_t \mathbf{I}),
\end{equation}
where $\{x_t\}_{t=1}^T$ represents the sequence of noisy states, $\beta_t \in (0, 1)$ is a variance schedule, and $x_0$ is the original data sample. This process can be expressed in closed form for any timestep $t$:
\begin{equation}
q(x_t | x_0) = \mathcal{N}(x_t; \sqrt{\bar{\alpha}_t} x_0, (1 - \bar{\alpha}_t) \mathbf{I}),
\end{equation}
where $\alpha_t = 1 - \beta_t$ and $\bar{\alpha}_t = \prod_{s=1}^t \alpha_s$. The generative process reverses this diffusion through a learned denoising model $\epsilon_\theta$ that predicts the noise component at each step:
\begin{equation}
p_\theta(x_{t-1} | x_t) = \mathcal{N}(x_{t-1}; \mu_\theta(x_t, t), \sigma_t^2 \mathbf{I}),
\end{equation}
where $\mu_\theta(x_t, t) = \frac{1}{\sqrt{\alpha_t}}\left(x_t - \frac{\beta_t}{\sqrt{1 - \bar{\alpha}_t}}\epsilon_\theta(x_t, t)\right)$. The model is then reparameterized and trained to minimize the simplified objective:
\begin{equation}
\mathcal{L} = \mathbb{E}_{t \sim [1,T], x_0 \sim q(x_0), \epsilon \sim \mathcal{N}(0, \mathbf{I})}\left[\|\epsilon - \epsilon_\theta(x_t, t)\|_2^2\right],
\end{equation}
which effectively trains the model to predict the noise that was added at each timestep. 
We can now alter the formulation once more to enable for the creation of conditional diffusion models that generate samples conditioned on input states. Specifically, we implement classifier-free guidance \cite{ho2022classifierfreediffusionguidance}, which provides a more stable training framework, better results, and general usage compared to the traditional classifier-guided \cite{dhariwal2021diffusionmodelsbeatgans} approaches. In classifier-free guidance, the model is trained with two objectives:
\begin{equation}
\mathcal{L}_{\text{CFG}} = \lambda\mathbb{E}\left[\|\epsilon - \epsilon_\theta(x_t, t, c)\|_2^2\right] + (1-\lambda)\mathbb{E}\left[\|\epsilon - \epsilon_\theta(x_t, t, \emptyset)\|_2^2\right],
\end{equation}
where $c$ represents the conditioning information, $\emptyset$ represents an unconditional sample, and $\lambda \sim \text{Bernoulli}(p)$. During inference, we interpolate between conditional and unconditional predictions:
\begin{equation}
\hat{\epsilon}_\theta(x_t, t, c) = \epsilon_\theta(x_t, t, \emptyset) + w \cdot (\epsilon_\theta(x_t, t, c) - \epsilon_\theta(x_t, t, \emptyset)),
\end{equation}
where $w$ is a guidance scale that controls how strictly the generation adheres to the conditioning. By adjusting $w$, we can balance between diversity (lower $w$) and fidelity (higher $w$) in our ensemble forecasts. 

Additionally, DDPMs discrete formulation can also be reformulated as a continuous process stochastic differential equation (SDE). Under this formulation we can make use ODE-based solvers like DPM-Solver \cite{lu2022dpmsolverfastodesolver} and DPM-Solver++ \cite{lu2023dpmsolverfastsolverguided} to accelerate our sampling process. The DPM-Solver++ formulation uses a semi-linear ODE:
\begin{equation}
\frac{dx}{dt} = f_1(t)x + f_2(t)s_\theta(x, t),
\end{equation}
where $s_\theta(x, t)$ is the learned score function, and $f_1(t)$ and $f_2(t)$ are time-dependent coefficients derived from the noise schedule. In this approach, we can use efficient high-order numerical methods such as the Runge-Kutta schemes \cite{tadmor2023rungekuttamethodsstable} to acheieve high-quality samples with as few as $T' = 10$ to $20$ function evaluations.

\section{Methodology}

\subsection{ERA5 Reanalysis Dataset}
ERA5 is the fifth generation atmospheric reanalysis dataset produced by the European Centre for Medium-Range Weather Forecasts (ECMWF), providing a comprehensive record of the global climate from 1950 to the present. The dataset is generated using the ECMWF's Integrated Forecast System (IFS) CY41R2, which assimilates observational data from satellites, weather stations, buoys, and other sources. The IFS model operates at a spatial resolution of approximately 31 km horizontally with 137 vertical levels, extending from the surface up to 0.01 hPa (approximately 80 km), enabling ERA5 to resolve atmospheric phenomena across multiple scales from global circulation patterns to mesoscale features.
The temporal resolution of ERA5 is hourly, although for this study we utilize data at six-hour intervals. ERA5 employs 4D-Var data assimilation techniques, which minimize a cost function measuring the difference between the model state and observations over a 12-hour window. This approach combines observations with short-range forecasts to produce a dynamically consistent estimate of the atmospheric state.
For our experiments, we utilize a specific subset of ERA5 variables at selected pressure levels as detailed in Table \ref{tab:dataset}. The atmospheric variables are sampled across 13 pressure levels from 1000 hPa to 50 hPa, capturing the vertical structure from near-surface conditions through the troposphere and into the lower stratosphere. The selected variables include both atmospheric parameters (geopotential, wind components, specific humidity, and temperature) and surface parameters (near-surface winds, 2-meter temperature, pressure measurements, water content, and precipitation).

\begin{table}[h]
\centering
\begin{tabular}{l p{3.5cm} p{6.8cm}}
\toprule
\textbf{Category} & \textbf{Variable} & \textbf{Description} \\
\midrule
Atmospheric & Geopotential & Height of pressure level (m$^2$s$^{-2}$) \\
& Wind components & Zonal, meridional, and vertical wind (m s$^{-1}$) \\
& Specific humidity & Mass of water vapor per unit mass of air (kg kg$^{-1}$) \\
& Temperature & Air temperature at pressure level (K) \\
\midrule
Surface & 10m wind (u, v) & Near-surface wind components (m s$^{-1}$) \\
& 2m temperature & Near-surface air temperature (K) \\
& Mean sea level pressure & Pressure reduced to sea level (hPa) \\
& Surface pressure & Pressure at surface (hPa) \\
& Total column water & Integrated water vapor in atmospheric column (kg m$^{-2}$) \\
& Total precipitation & Accumulated precipitation over 6 hours (m) \\
\bottomrule
\end{tabular}
\caption{ERA5 variables used in this study}
\label{tab:dataset}
\end{table}

We structure our data as a tensor $x_t \in \mathbb{R}^{v \times h \times w}$, where $v$ denotes the number of variables and $h, w$ represent the spatial dimensions. Beyond the standard atmospheric and surface variables from ERA5, we augment our input with 11 additional forcing variables and constants (detailed in Table \ref{tab:forcings}). These supplementary variables provide context while remaining state-independent, as they can be deterministically computed for any spatiotemporal coordinate. By concatenating these forcing variables along the first dimension, we obtain an expanded input tensor of size $[v+f, h \times w]$, where $f=11$ represents the number of auxiliary variables. It is important to note the asymmetry in our processing pipeline: while the neural forecasting model $G_\theta$ consumes the full augmented input $[v+f, h, w]$, it generates predictions only for the ERA5 variables, yielding outputs of size $[v, h, w]$. Contrary to this, the diffusion-based condition perturbation model $\epsilon_\theta$ operates exclusively on the ERA5 variables of size $[v, h, w]$, as the forcing variables remain constant and deterministic throughout the prediction process.

\begin{table}[h]
\centering
\begin{tabular}{l l l}
\toprule
\textbf{Category} & \textbf{Variables} & \textbf{Description} \\
\midrule
Temporal & $\sin(\text{time of day})$ & Diurnal cycle representation \\
Forcings & $\cos(\text{time of day})$ & using circular encoding \\
& $\sin(\text{year progress})$ & Annual cycle representation \\
& $\cos(\text{year progress})$ & using circular encoding \\
\midrule
Energy   & TOA radiation & Top-of-atmosphere radiation flux \\
Input    &              & integrated over preceding hour \\
\midrule
Geographical & Latitude & Normalized spatial coordinates \\
Information  & Longitude & and surface characteristics \\
& Land-sea mask & Binary indicator (1 = land, 0 = ocean) \\
& Topography & Surface elevation data \\
& Leaf area index & Vegetation density measure \\
& Albedo & Surface reflectivity coefficient \\
\bottomrule
\end{tabular}
\caption{Forcing variables and constants incorporated into the input data tensor.}
\label{tab:forcings}
\end{table}

Prior to feeding our data through either the neural forecasting model $G_\theta$ or the diffusion-based condition perturbation model $\epsilon_\theta$, we implement a pre-processing normalization procedure for each variable, across its spatial dimensions. Specifically, we compute and store the mean $\mu_v$ and standard deviation $\sigma_v$ for each variable. These statistics are subsequently used to denormalize the outputs after model processing. While this normalization approach inevitably discards some distributional information, it serves a critical purpose: preventing the learning process from being disproportionately influenced by variables with larger magnitudes or wider ranges. This ensures that the model weights reflect the relative importance of physical relationships rather than numerical scale differences. The normalization procedure is provided in \ref{appendix:algos}.

\begin{figure}[h!]
    \centering
    \begin{subfigure}[b]{0.48\textwidth}
        \centering
        \includegraphics[width=\textwidth]{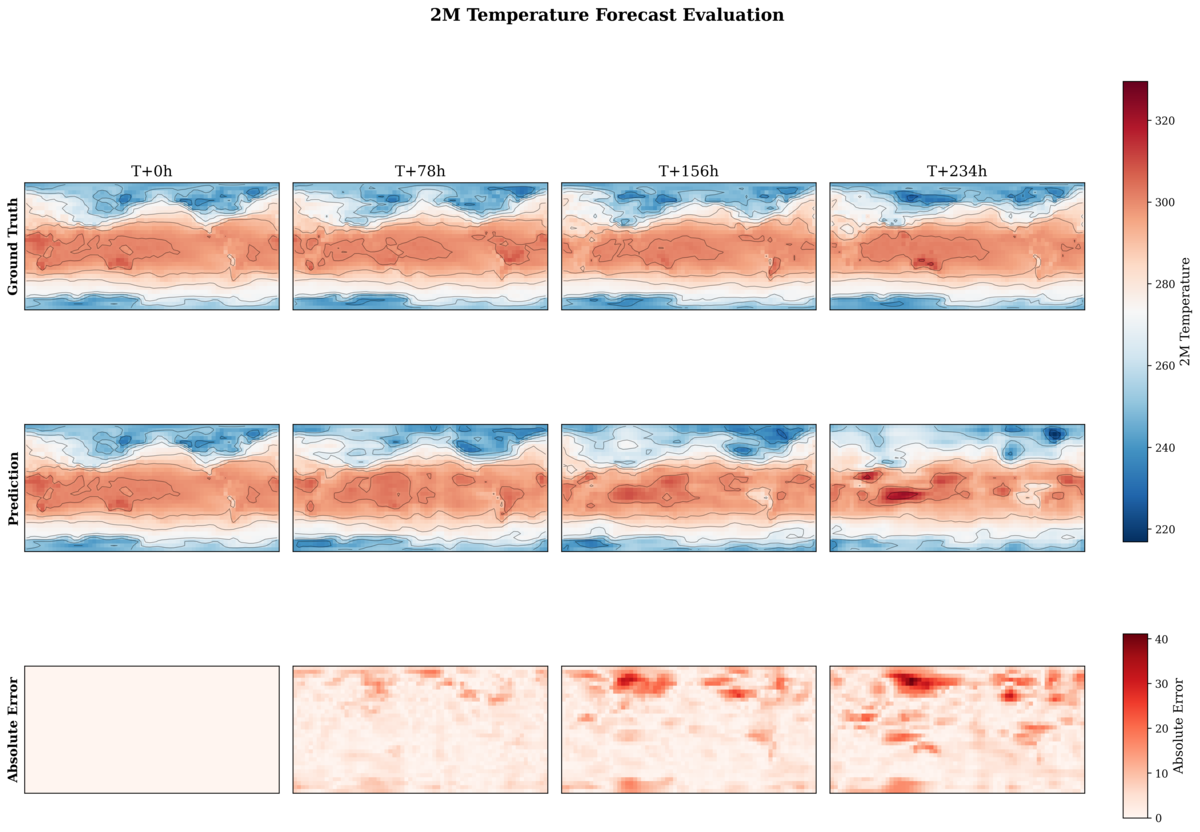}
        \label{fig:deter_temp}
    \end{subfigure}
    \hfill
    \begin{subfigure}[b]{0.48\textwidth}
        \centering
        \includegraphics[width=\textwidth]{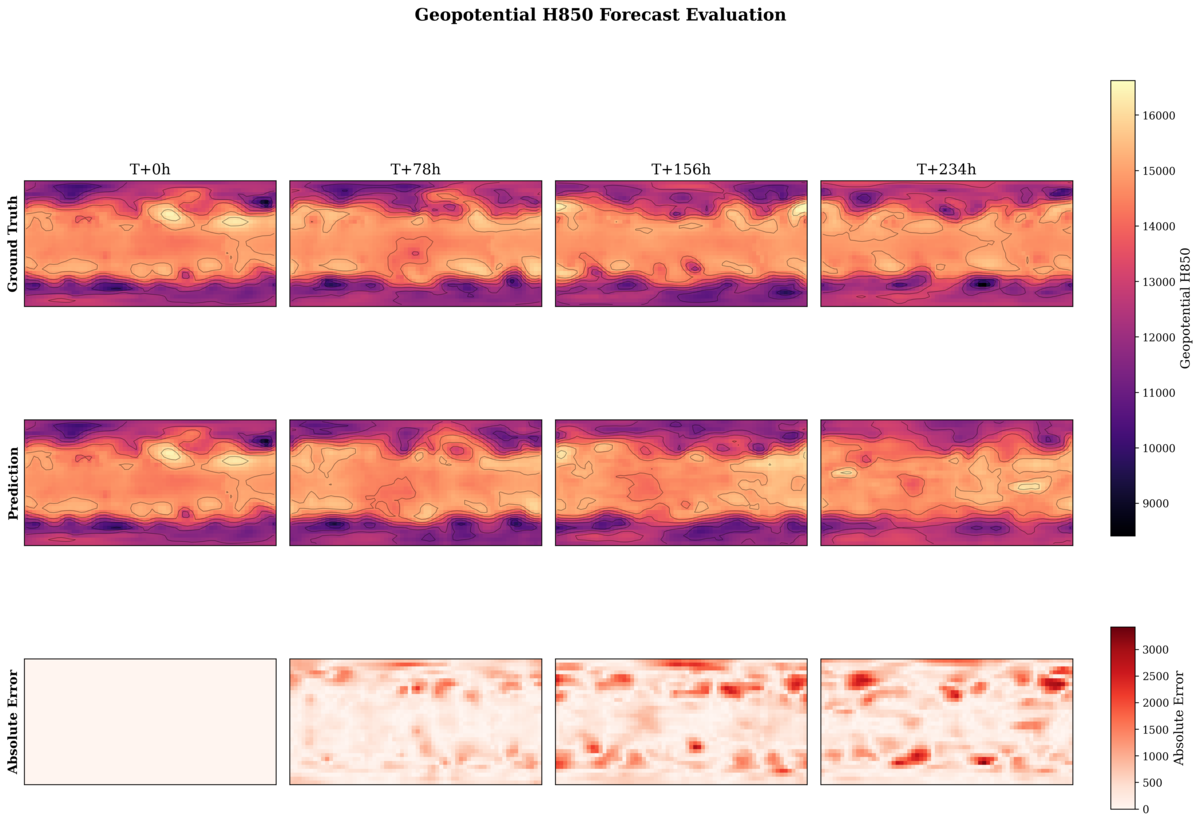}
        \label{fig:deter_geo}
    \end{subfigure}
    
    \vspace{0.5cm}
    
    \begin{subfigure}[b]{0.48\textwidth}
        \centering
        \includegraphics[width=\textwidth]{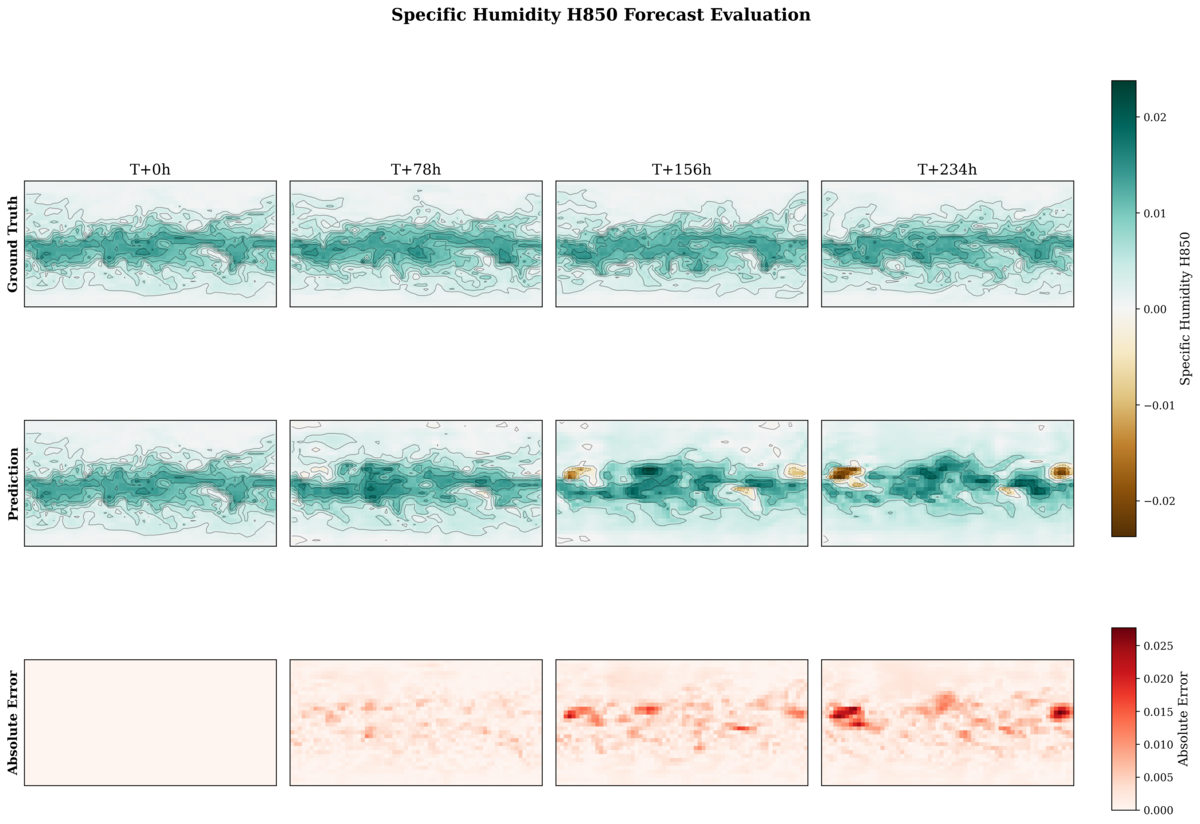}
        \label{fig:deter_humid}
    \end{subfigure}
    \hfill
    \begin{subfigure}[b]{0.48\textwidth}
        \centering
        \includegraphics[width=\textwidth]{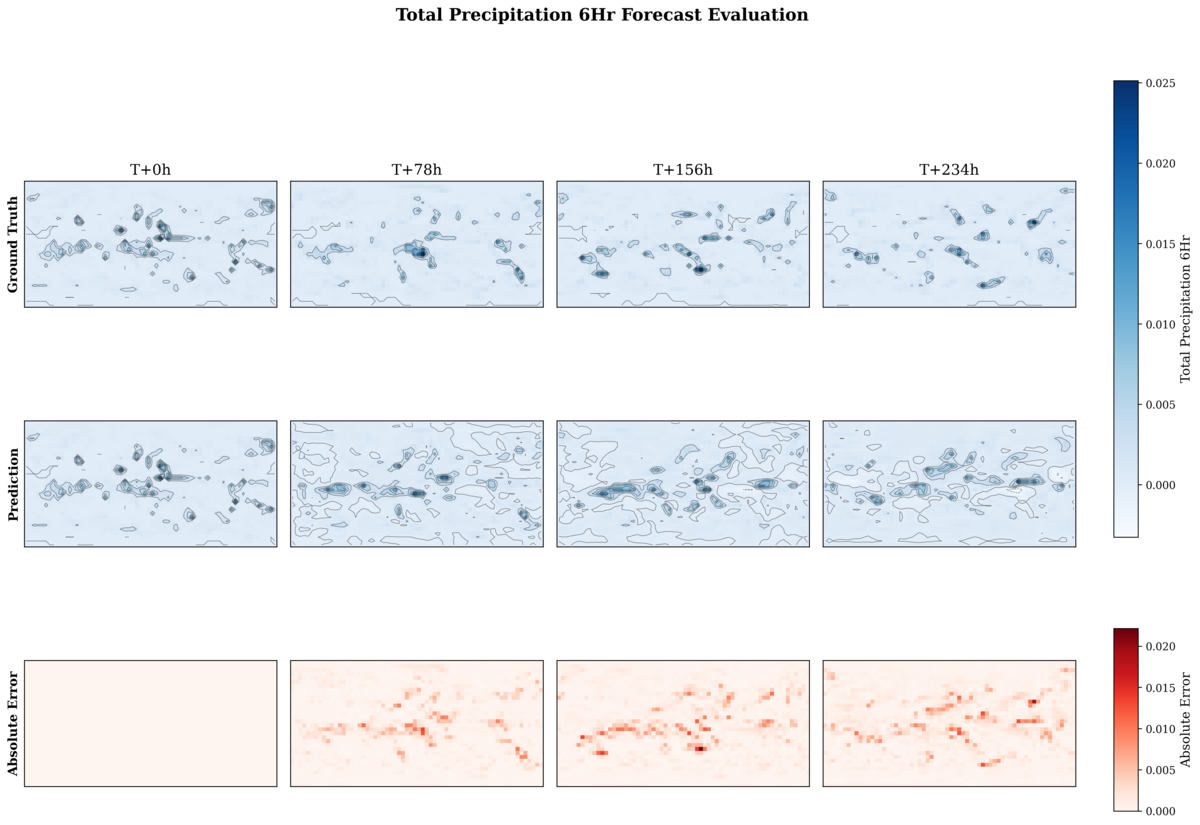}
        \label{fig:deter_rain}
    \end{subfigure}
    \caption{Snapshots of meteorological variables from deterministic forecast: (a) 2m temperature, (b) geopotential height at 850 hPa, (c) specific humidity at 850 hPa, and (d) 6-hour total precipitation.}
    \label{fig:deterministic_variables}
\end{figure}

\subsection{Models}
The forecasting model $G_\phi$ predicts the next state $x_{t+1}$ given the current state $x_t$ and is trained using the mean squared error (MSE) objective. The diffusion-based perturbation model $\epsilon_\theta$ functions as a stochastic reconstruction mechanism: conditioned on $x_t$ or $\hat{x}_t$, it generates a sample from the full diffusion process of $x_t$ or $\hat{x}_t$. While our framework does not impose architectural constraints, we adopt an attention-based U-Net for $G_\phi$ due to its ability to efficiently capture both local and global dependencies. The network consists of four downsampling and four upsampling blocks, incorporating wide ResNet layers and spatial attention layers at resolutions $32 \times 16$, $16 \times 8$, and $8 \times 4$. The perturbation model $\epsilon_\theta$ employs the same architecture but includes necessary time embedding layers at each stage, as well as an extra attention layer at the $64 \times 32$ resolution. We provide complete algorithms for both training processes in ~\ref{appendix:algos}.

During inference, we combine the two models to obtain a distribution of ensemble members given a ground truth observation. Consider an initial ground truth observation $x_0$. In the deterministic case, to forecast for $N$ timesteps, we autoregressively apply the forecasting model, denoted by:
\begin{equation}
    x_{t+1} = G_\phi(x_t), \quad \forall t \in \{0, \dots, N-1\}.
\end{equation}
This formulation yields a single trajectory from the ground truth. With access to the perturbation model, we introduce a stochastic component before each autoregression step to perturb each state. To obtain an ensemble of trajectories, we generate $B$ perturbed versions of the initial observation and propagate them forward in parallel:
\begin{equation}
    x_{t+1}^{(b)} = G_\phi(\tilde{x}_t^{(b)}), \quad \tilde{x}_t^{(b)} = \epsilon_\theta(x_t^{(b)}), \quad \forall b \in \{1, \dots, B\}.
\end{equation}
Here, $\tilde{x}_t^{(b)}$ represents the perturbed state obtained from the diffusion model, and $x_{t+1}^{(b)}$ represents the corresponding forecasted state for each ensemble member. This procedure allows us to directly control exploration through guidance term of the conditional diffusion model. Additionally, it enables random walks via an iterative process of applying:
\begin{equation} 
    \tilde{x}_{t, k}^{(b)} = \epsilon_\theta(x_{t, k-1}^{(b)}), \quad \forall k \in \{1, \dots, K\},
\end{equation}
where $K$ denotes the number of perturbation steps taken per state. This iterative application of $\epsilon_\theta$ effectively simulates multi-step stochastic transitions, rivaling the additional exploration of the guidance term. 

\begin{figure}[h!]
    \centering
    \begin{subfigure}[b]{0.48\textwidth}
        \centering
        \includegraphics[width=\textwidth]{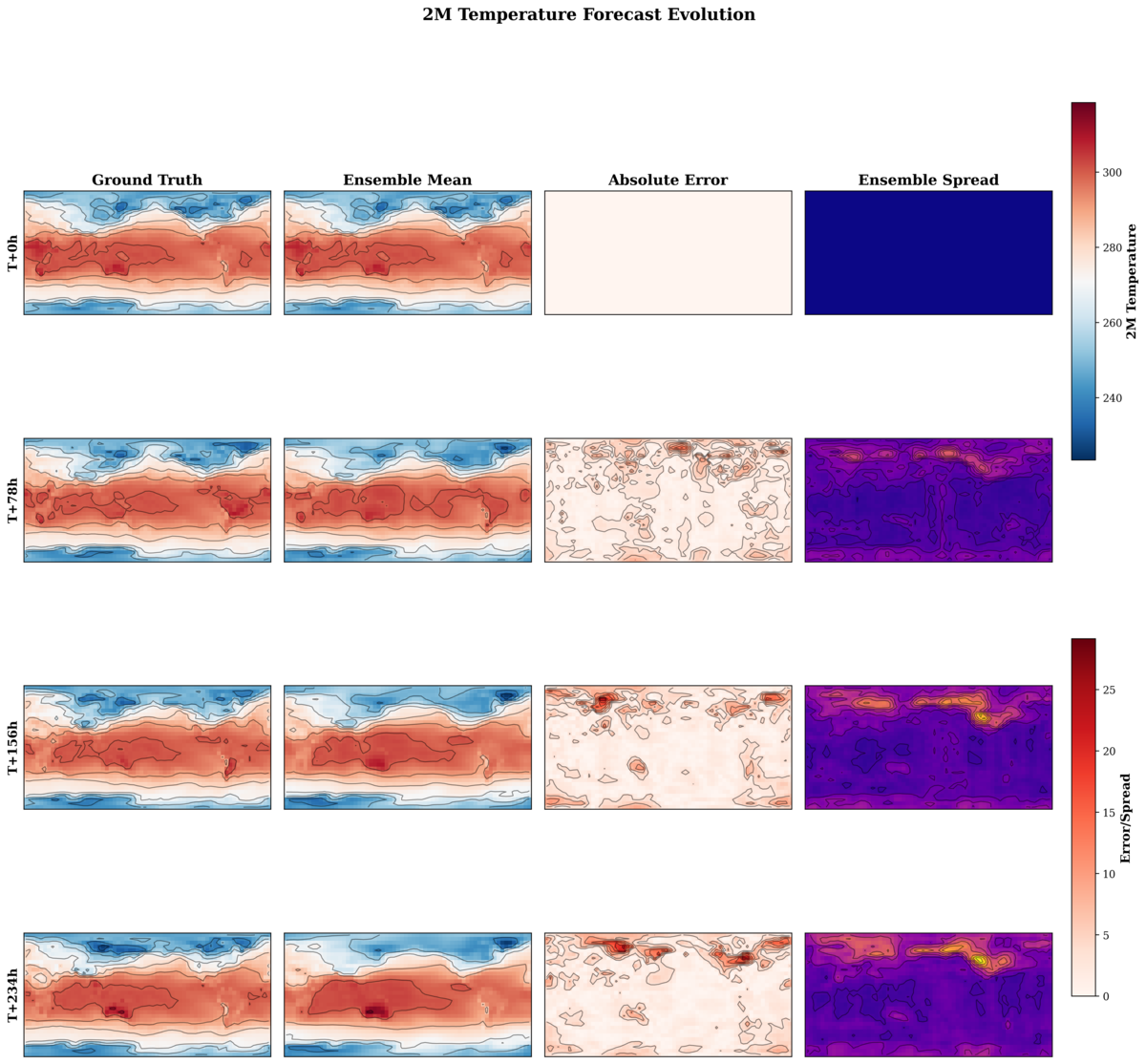}
        \label{fig:compare_2m}
    \end{subfigure}
    \hfill
    \begin{subfigure}[b]{0.48\textwidth}
        \centering
        \includegraphics[width=\textwidth]{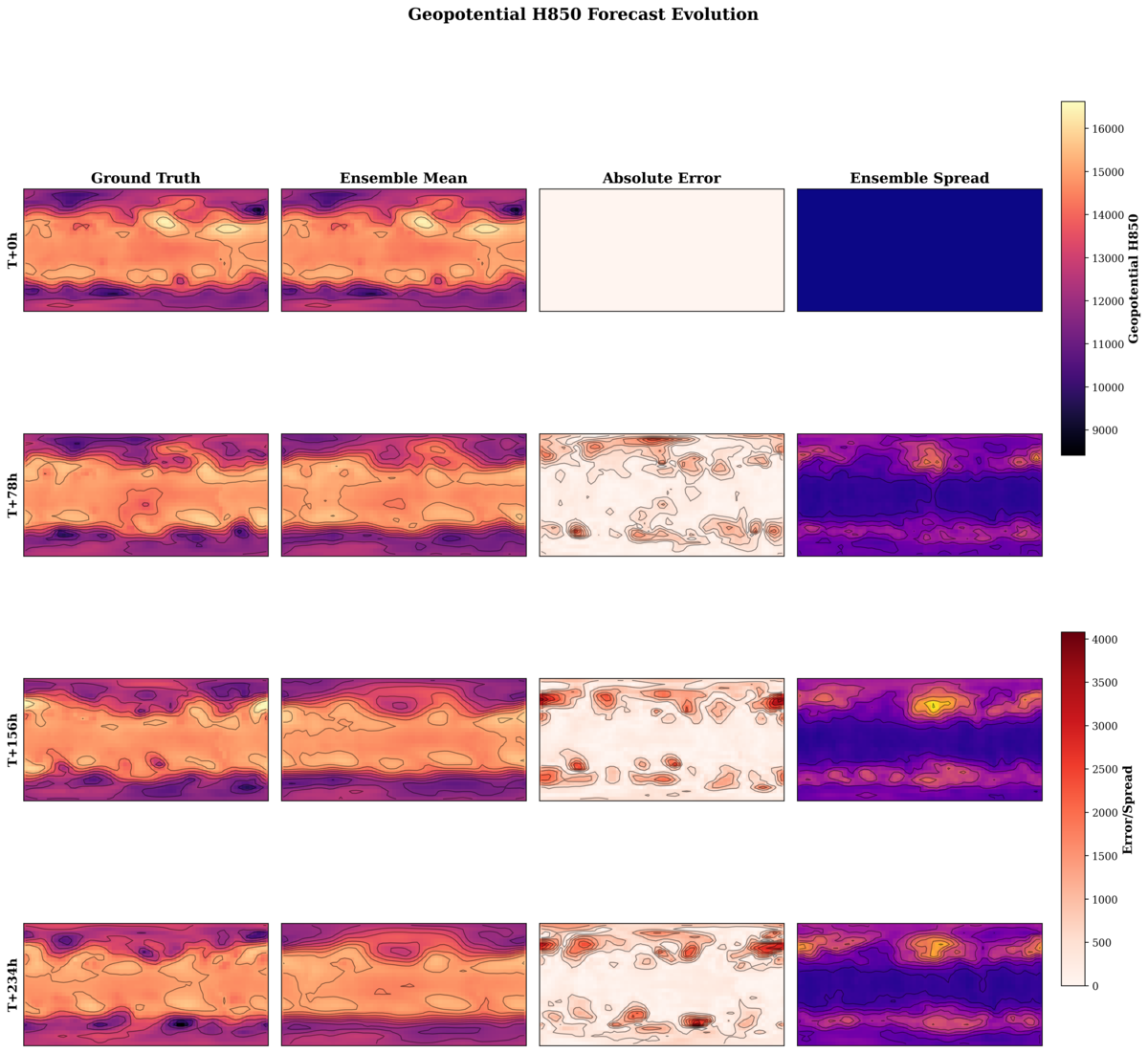}
        \label{fig:compare_geo}
    \end{subfigure}
    
    \vspace{0.5cm}
    
    \begin{subfigure}[b]{0.48\textwidth}
        \centering
        \includegraphics[width=\textwidth]{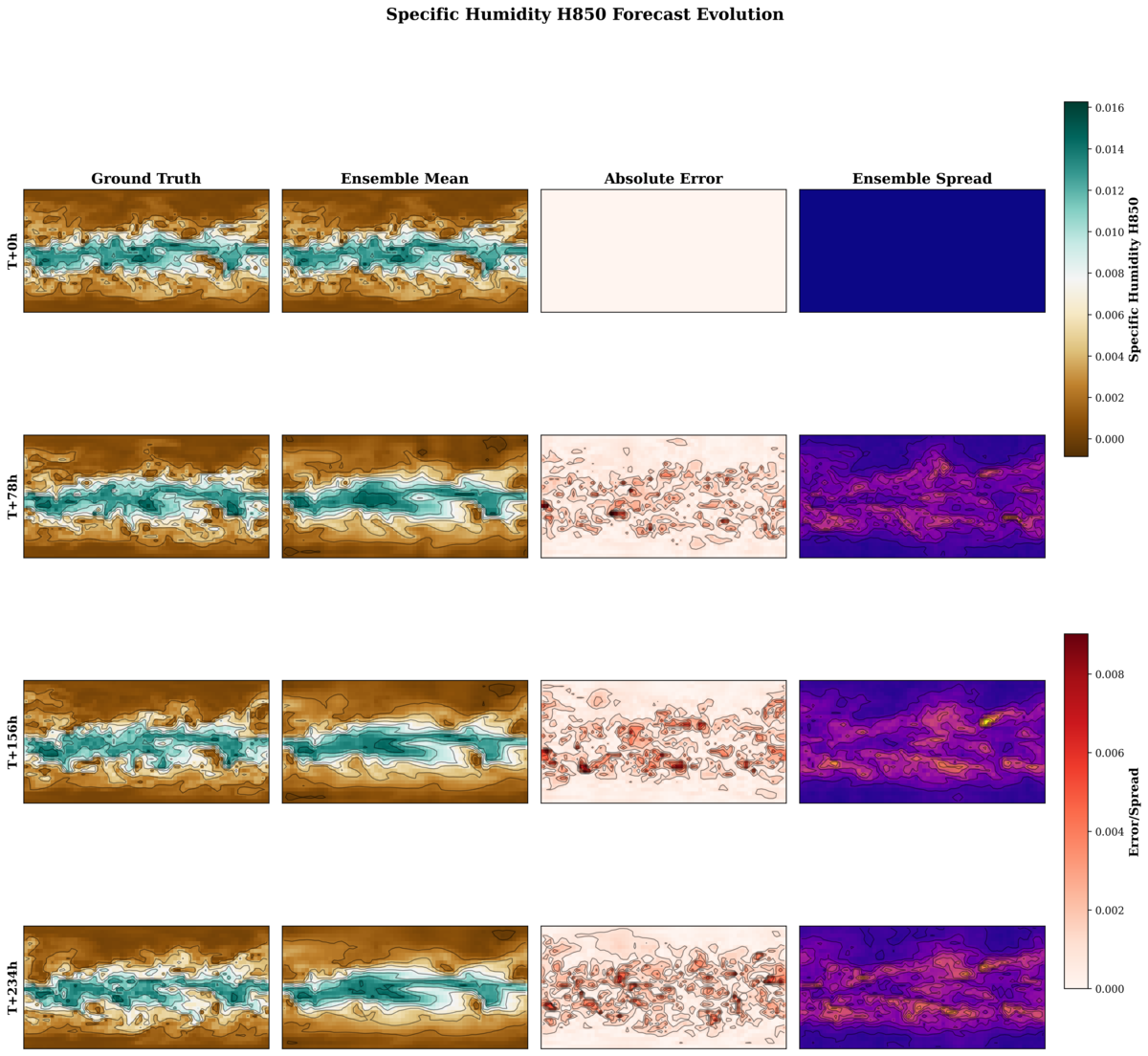}
        \label{fig:compare_humid}
    \end{subfigure}
    \hfill
    \begin{subfigure}[b]{0.48\textwidth}
        \centering
        \includegraphics[width=\textwidth]{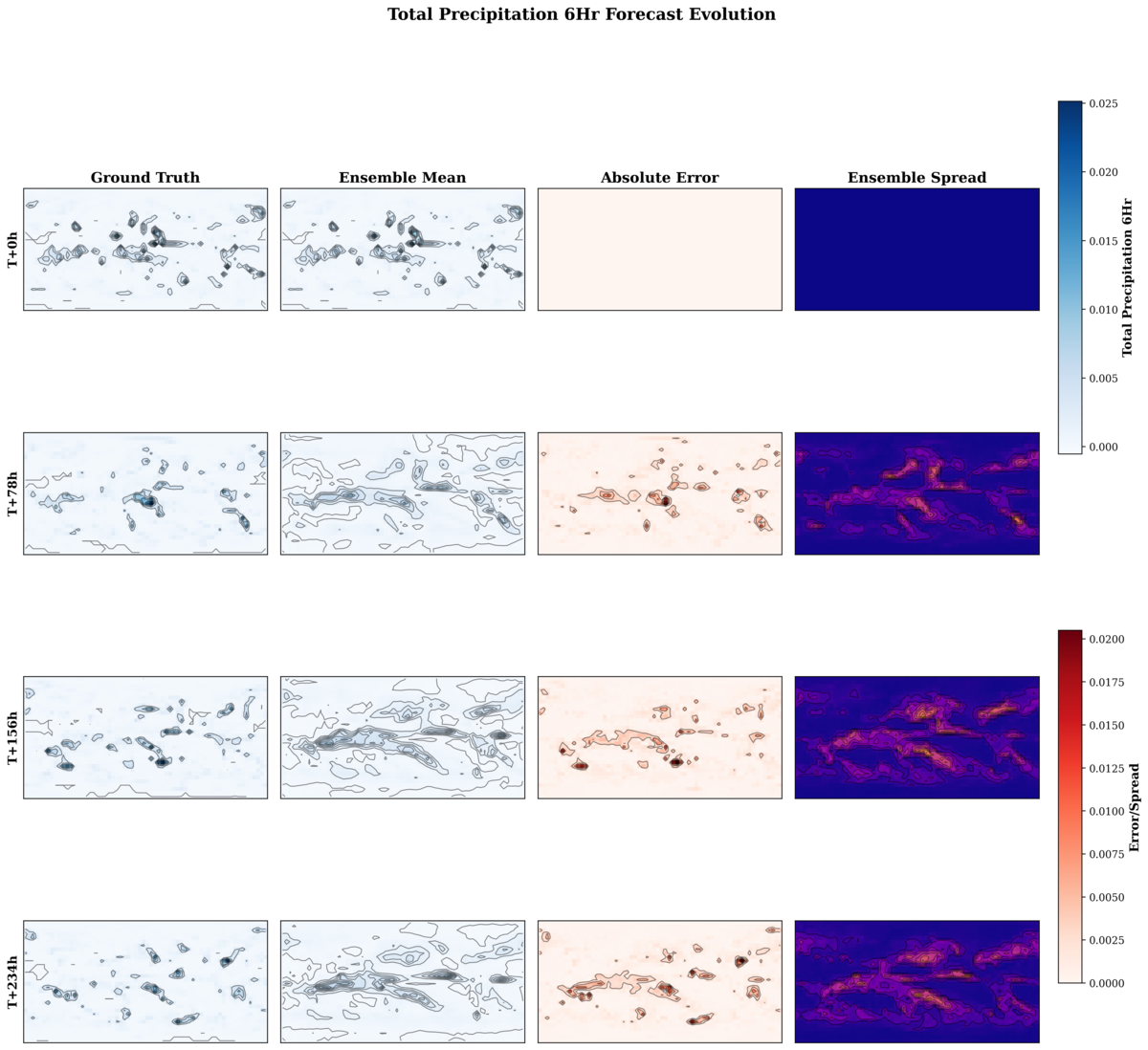}
        \label{fig:compare_rain}
    \end{subfigure}
    \caption{Evaluation of diffusion-based ensemble forecasts showing mean prediction, ground truth, forecast error, and ensemble spread for: (a) 2-meter temperature, (b) 850 hPa geopotential height, (c) 850 hPa specific humidity, and (d) 6-hour precipitation.}
    \label{fig:comparison_variables}
\end{figure}

\begin{figure}[h!]
    \centering
    \begin{subfigure}[b]{0.48\textwidth}
        \centering
        \includegraphics[width=\textwidth]{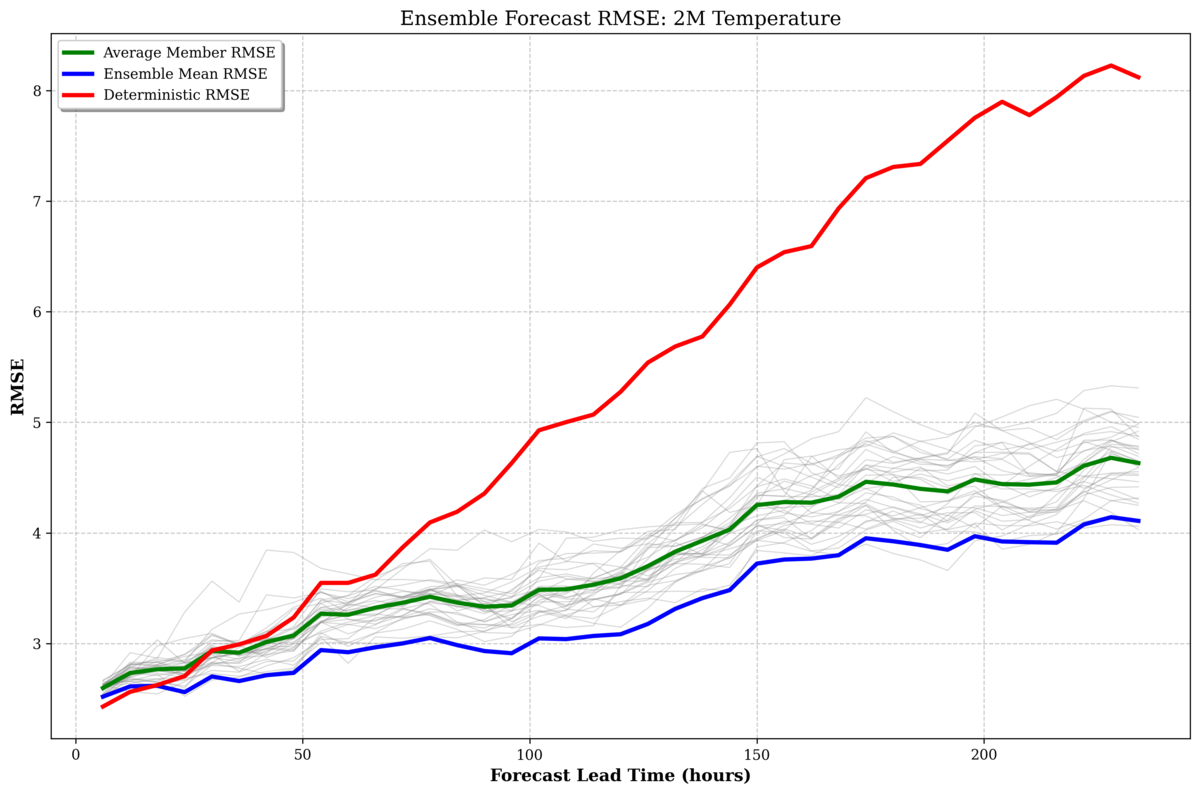}
        \label{fig:traj_temp}
    \end{subfigure}
    \hfill
    \begin{subfigure}[b]{0.48\textwidth}
        \centering
        \includegraphics[width=\textwidth]{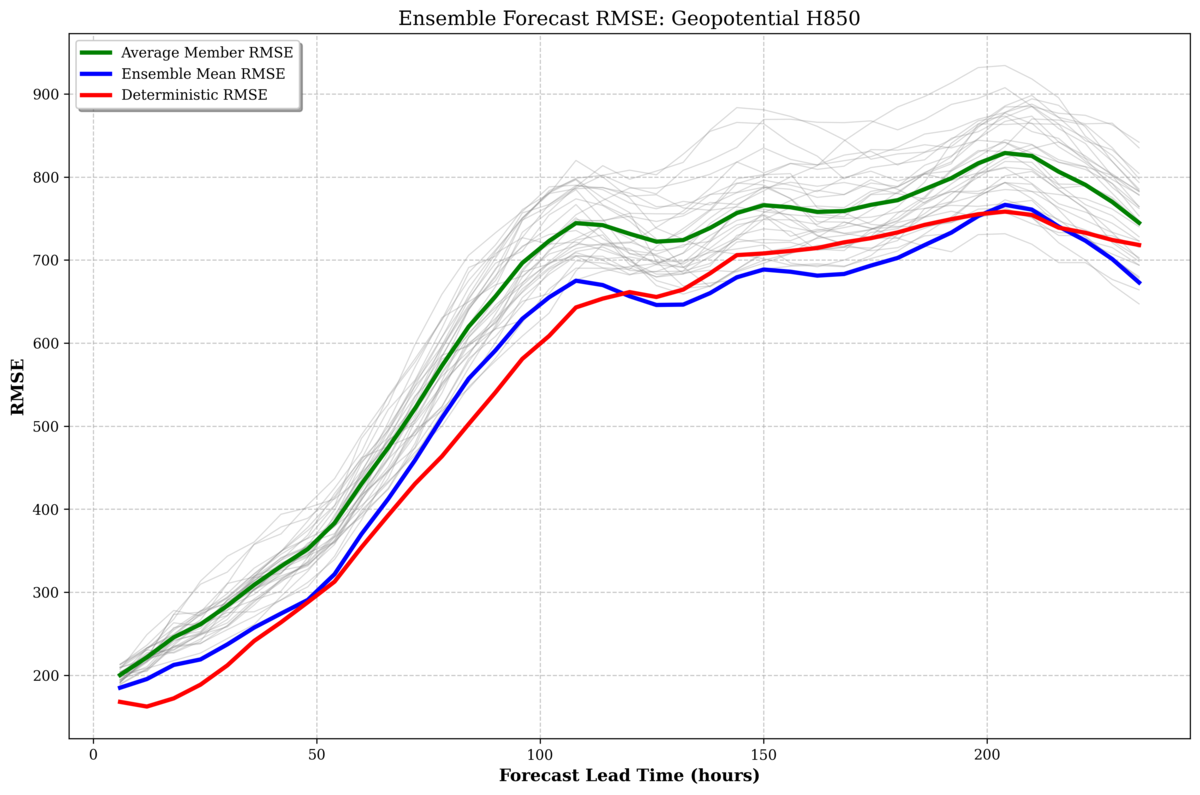}
        \label{fig:traj_geo}
    \end{subfigure}
    
    \vspace{0.5cm}
    
    \begin{subfigure}[b]{0.48\textwidth}
        \centering
        \includegraphics[width=\textwidth]{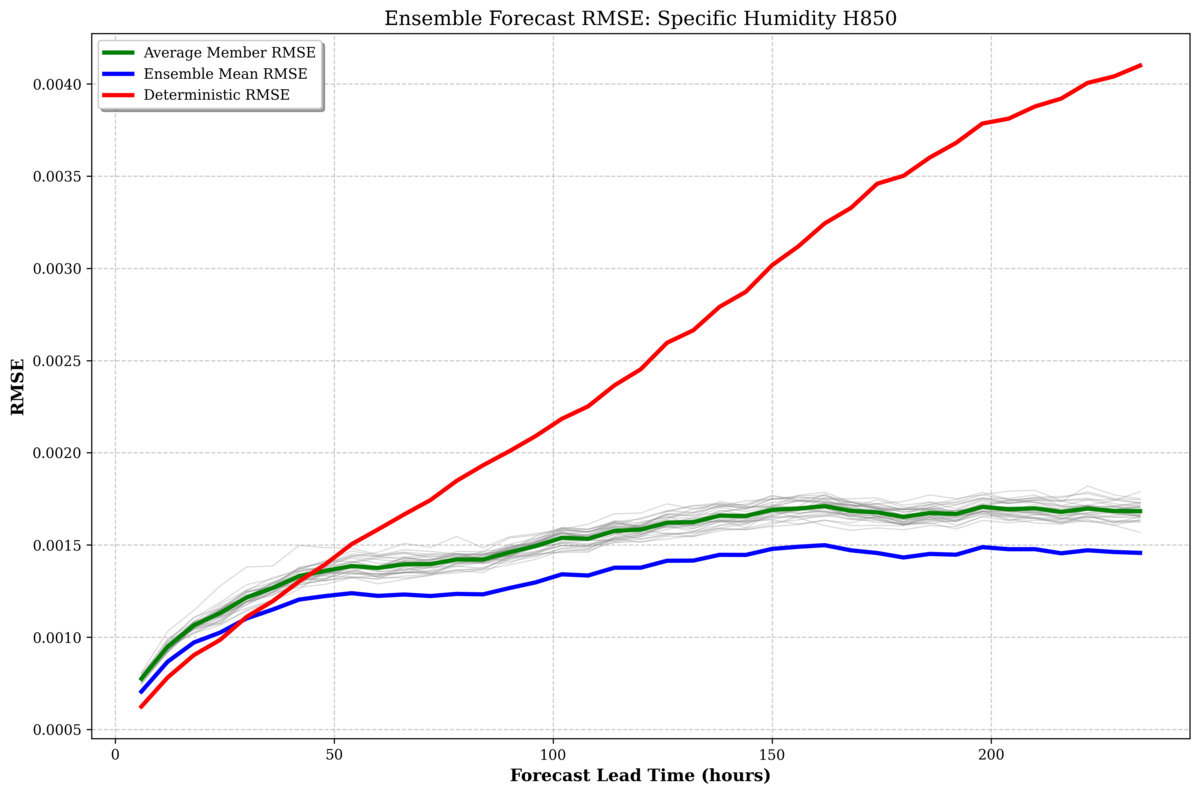}
        \label{fig:traj_humid}
    \end{subfigure}
    \hfill
    \begin{subfigure}[b]{0.48\textwidth}
        \centering
        \includegraphics[width=\textwidth]{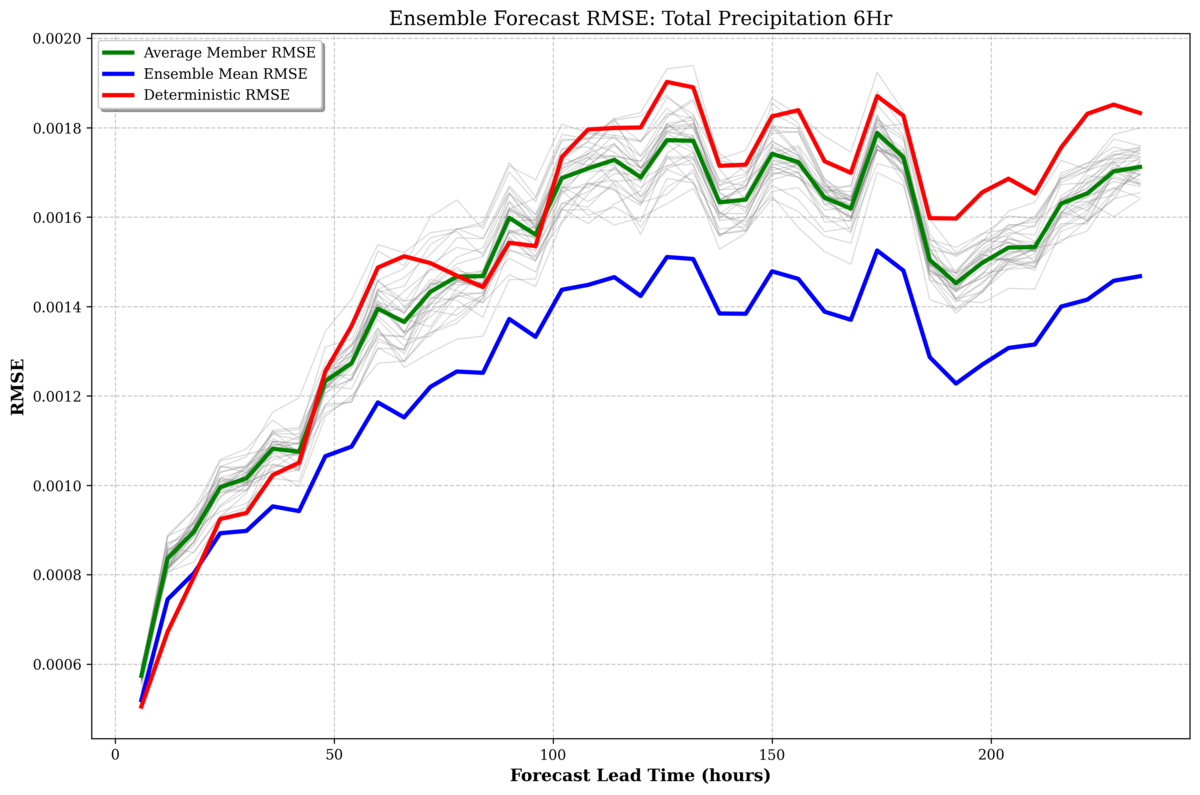}
        \label{fig:traj_rain}
    \end{subfigure}
    \caption{Comparison of ensemble forecast error metrics: individual ensemble members' RMSE versus ground truth (gray lines), mean RMSE across all ensemble members (green line), and RMSE of the ensemble mean prediction versus ground truth (blue line) for: (a) 2-meter temperature, (b) 850 hPa geopotential height, (c) 850 hPa specific humidity, and (d) 6-hour accumulated precipitation.}
    \label{fig:rmse_comparison}
\end{figure}

\subsection{Evaluation Metrics}
This study uses four quantitative evaluation metrics: Continuous Ranked Probability Score (CRPS), Energy Score, Root Mean Squared Error (RMSE), and Spread Correlation (SC). We also consider qualitative differences, as quantifying fidelity across 85 variables is too challenging for metrics such as SSIM and PNR. Among these, RMSE is the sole metric that allows comparison between stochastic and deterministic predictions. We formally define CRPS as:
\begin{equation}
    \text{CRPS}(x, \{x^{(b)}\}_{b=1}^{B}) = \frac{1}{B} \sum_{b=1}^{B} |x^{(b)} - x| - \frac{1}{2B^2} \sum_{b=1}^{B} \sum_{b'=1}^{B} |x^{(b)} - x^{(b')}|,
\end{equation}
where \( x \) is the observed ground truth value, and \( \{x^{(b)}\}_{b=1}^{B} \) represents the ensemble of predictions. CRPS generalizes the mean absolute error (MAE) to probabilistic forecasts, measuring both accuracy and sharpness. Lower CRPS values indicate better probabilistic calibration.
The Energy Score evaluates ensemble forecasts similarly to CRPS and is defined as:
\begin{equation}
    \text{ES}(\{x^{(b)}\}_{b=1}^{B}, x) = \frac{1}{B} \sum_{b=1}^{B} \| x^{(b)} - x \|_2 - \frac{1}{2B^2} \sum_{b=1}^{B} \sum_{b'=1}^{B} \| x^{(b)} - x^{(b')} \|_2,
\end{equation}
where \( \{x^{(b)}\}_{b=1}^{B} \) represents an ensemble of predictions, and \( x \) is the observed ground truth. Compared to CRPS, the Energy Score uses the Euclidean norm \( \|\cdot \|_2 \) instead of the Manhattan norm. A lower Energy Score implies better sharpness and accuracy of the ensemble.
RMSE measures the pointwise error of single forecasts and is defined as:
\begin{equation}
    \text{RMSE}(x, \hat{x}) = \sqrt{\frac{1}{n} \sum_{i=1}^{n} (x_i - \hat{x}_i)^2},
\end{equation}
where \( x_i \) represents the ground truth values, and \( \hat{x}_i \) represents the predicted values. Metrics such as RMSE, MAE, and MSE are widely used for measuring prediction error, as they provide a single scalar value to quantify discrepancies in high-dimensional spaces. SC measures the distance between the error and pointwise standard deviation and is defined as:
\begin{equation}
    \text{SC}(\{x^{(b)}\}_{b=1}^{B}, x, \hat{x}) =  \left\| \left( \hat{x} - x \right) - \sqrt{\frac{1}{B} \sum_{b=1}^{B} \left( x^{(b)} - \bar{x} \right)^2} \right\|_2
\end{equation}
where $\bar{x}$ is the ensemble mean and $x^{(b)}$ are ensemble members. SC provides a quantitative measure of how well the ensemble represents the actual error. It compares the error of the mean ensemble prediction to the spread (or variability) within the ensemble members. A lower SC value (relative to the units) indicates that the ensemble accurately reflects the error and that the spread of the ensemble matches the true variability of the system, leading to a more reliable uncertainty estimate.

\begin{table}[h]
    \centering
    \setlength{\tabcolsep}{3pt} 
    \scriptsize 
    \begin{tabular}{l S S S S | S S S S }
        \toprule
        & \multicolumn{4}{c|}{\textbf{Lead Time: 60h}} & \multicolumn{4}{c}{\textbf{Lead Time: 120h}} \\
        \cmidrule(lr){2-5} \cmidrule(lr){6-9}
        & \multicolumn{1}{c}{Energy} & \multicolumn{1}{c}{CRPS} & \multicolumn{1}{c}{RMSE} & \multicolumn{1}{c|}{Spread} & \multicolumn{1}{c}{Energy} & \multicolumn{1}{c}{CRPS} & \multicolumn{1}{c}{RMSE} & \multicolumn{1}{c}{Spread} \\
        \midrule
        Deterministic & {-} & {-} & \num{3.55} & {-} & {-} & {-} & \num{5.28} & {-} \\
        Diffusion[0.3, 1] & \num{2.66} & \num{1.80} & \num{5.02} & \num{129.17} & \num{2.84} & \num{1.97} & \num{5.30} & \num{127.65} \\
        Diffusion[0.5, 1] & \textbf{\num{1.89}} & \textbf{\num{1.49}} & \textbf{\num{2.88}} & \textbf{\num{86.88}} & \textbf{\num{1.62}} & \textbf{\num{1.29}} & \textbf{\num{2.99}} & \textbf{\num{84.78}} \\
        Diffusion[0.7, 1] & \num{2.58} & \num{1.97} & \num{3.04} & \num{114.75} & \num{2.73} & \num{2.00} & \num{3.53} & \num{125.70} \\
        Diffusion[1.0, 1] & \num{3.42} & \num{2.57} & \num{3.58} & \num{153.99} & \num{5.09} & \num{3.55} & \num{5.37} & \num{229.83} \\
        \midrule
        & \multicolumn{4}{c|}{\textbf{Lead Time: 180h}} & \multicolumn{4}{c}{\textbf{Lead Time: 234h}} \\
        \cmidrule(lr){2-5} \cmidrule(lr){6-9}
        & \multicolumn{1}{c}{Energy} & \multicolumn{1}{c}{CRPS} & \multicolumn{1}{c}{RMSE} & \multicolumn{1}{c|}{Spread} & \multicolumn{1}{c}{Energy} & \multicolumn{1}{c}{CRPS} & \multicolumn{1}{c}{RMSE} & \multicolumn{1}{c}{Spread} \\
        \midrule
        Deterministic & {-} & {-} & \num{7.31} & {-} & {-} & {-} & \num{8.12} & {-} \\
        Diffusion[0.3, 1] & \num{2.92} & \num{2.21} & \num{5.64} & \num{139.36} & \num{4.03} & \num{2.93} & \num{6.49} & \num{157.76} \\
        Diffusion[0.5, 1] & \textbf{\num{2.28}} & \textbf{\num{1.51}} & \textbf{\num{3.78}} & \textbf{\num{114.75}} & \textbf{\num{2.42}} & \textbf{\num{1.61}} & \textbf{\num{4.03}} & \textbf{\num{118.24}} \\
        Diffusion[0.7, 1] & \num{3.32} & \num{2.17} & \num{4.52} & \num{152.88} & \num{3.12} & \num{1.97} & \num{4.56} & \num{148.90} \\
        Diffusion[1.0, 1] & \num{6.98} & \num{4.71} & \num{7.50} & \num{315.45} & \num{8.44} & \num{5.73} & \num{9.29} & \num{379.15} \\
        \bottomrule
    \end{tabular}
    \caption{2M Temperature forecast metrics across different lead times.}
    \label{tab:2m_temperature_metrics_guidance}
\end{table}

\subsection{Experimental Setup}
In this study, we assume access to limited computational resources. Our training is distributed across two NVIDIA RTX A5000 GPUs, each equipped with 24 GB of VRAM. Given these constraints, we opted for both the lower-resolution dataset and a more computationally efficient model. The deterministic model, $G_\phi$, was trained for 80 epochs with a batch size of 64, while the diffusion model, $\epsilon_\theta$, was trained for 300 epochs with a batch size of 32. Both models reached log-scale convergence. Inference was conducted within the same computational environment. We did not perform hyperparameter tuning and instead relied on the default settings of each package. In total, training took roughly 2 weeks, but could easily be accelerated and scaled with better hardware.

\section{Results}
In this section, we discuss three primary findings. First, we find that our diffusion-perturbation model produces valid ensembles that correlate well with the error of the mean ensemble prediction. Next, we explore the quantitative improvements over the deterministic prediction using the mean ensemble, along with other probabilistic metrics. Following this, we interpret the results of our two exploratory extensions: iterative walks and guidance terms. Finally, we observe that our method sometimes acts as a predictor-corrector, leading to individual ensemble members outperforming the deterministic prediction in certain situations. We do evaluation with the same start time of January 12th, 12 AM until January 22th, 12 AM, a 240 hour lead time.

\begin{figure}[h!]
    \centering
    \begin{subfigure}[b]{0.48\textwidth}
        \centering
        \includegraphics[width=\textwidth]{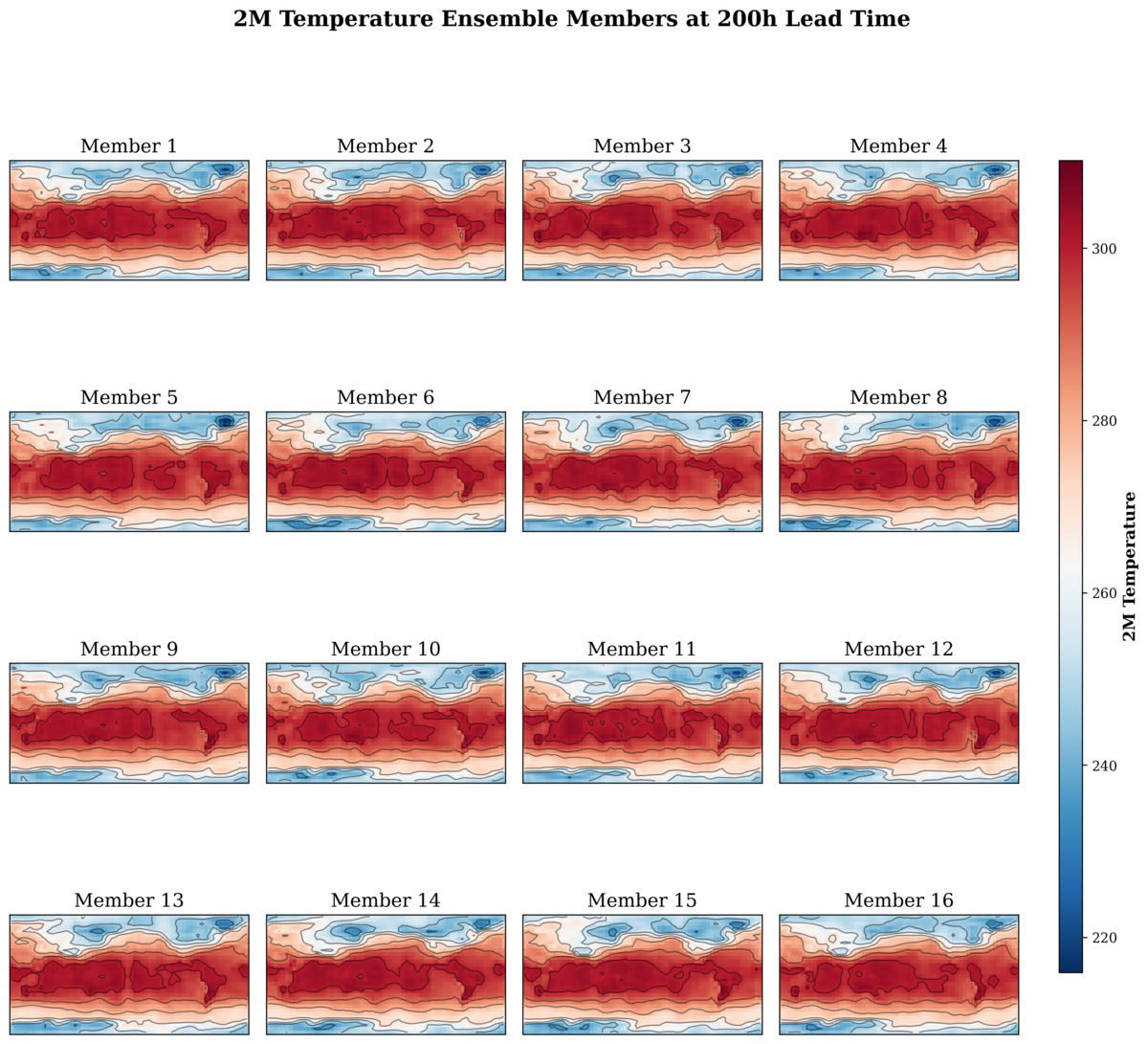}
        \label{fig:members_temp}
    \end{subfigure}
    \hfill
    \begin{subfigure}[b]{0.48\textwidth}
        \centering
        \includegraphics[width=\textwidth]{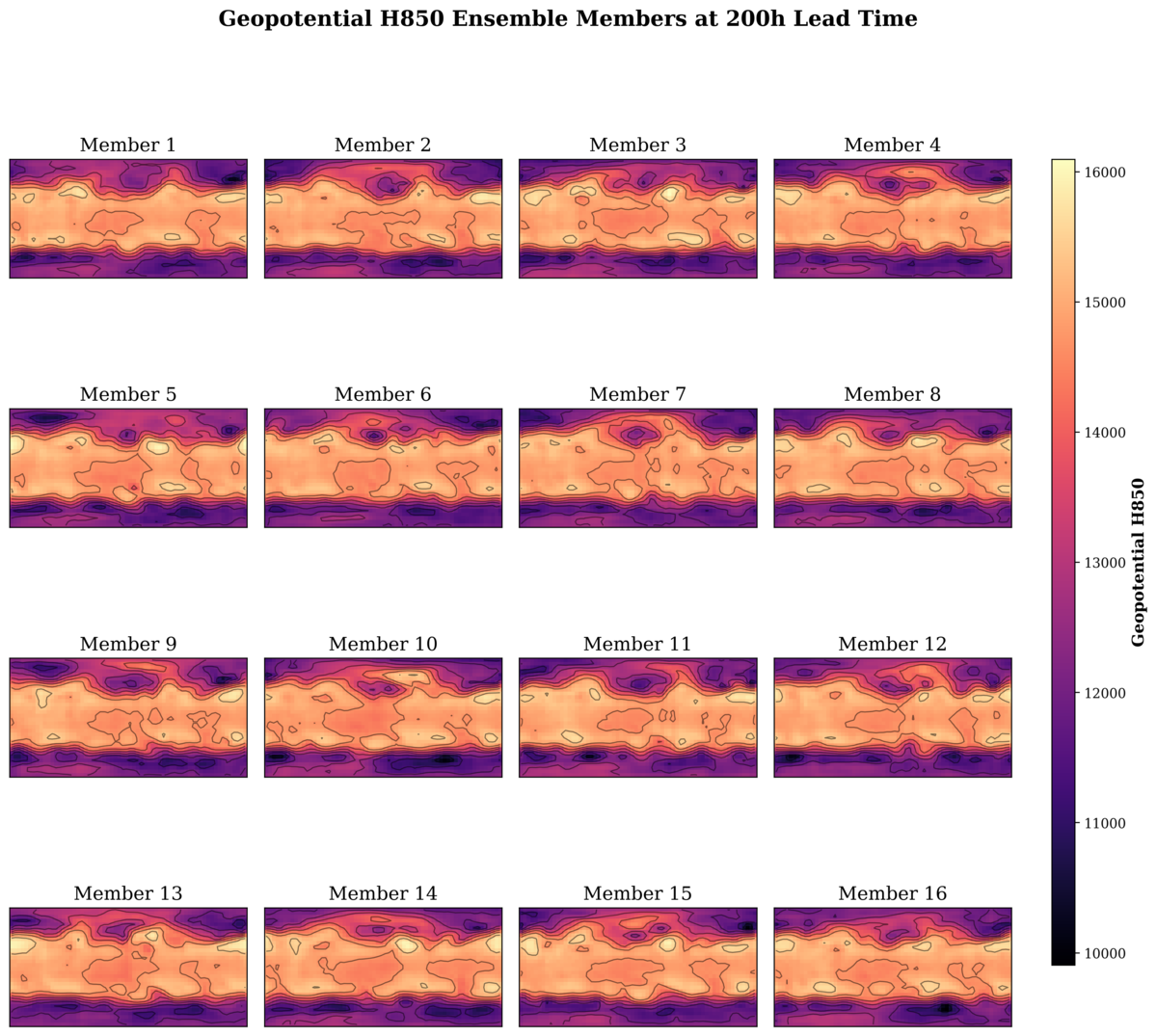}
        \label{fig:members_geo}
    \end{subfigure}
    
    \vspace{0.5cm}
    
    \begin{subfigure}[b]{0.48\textwidth}
        \centering
        \includegraphics[width=\textwidth]{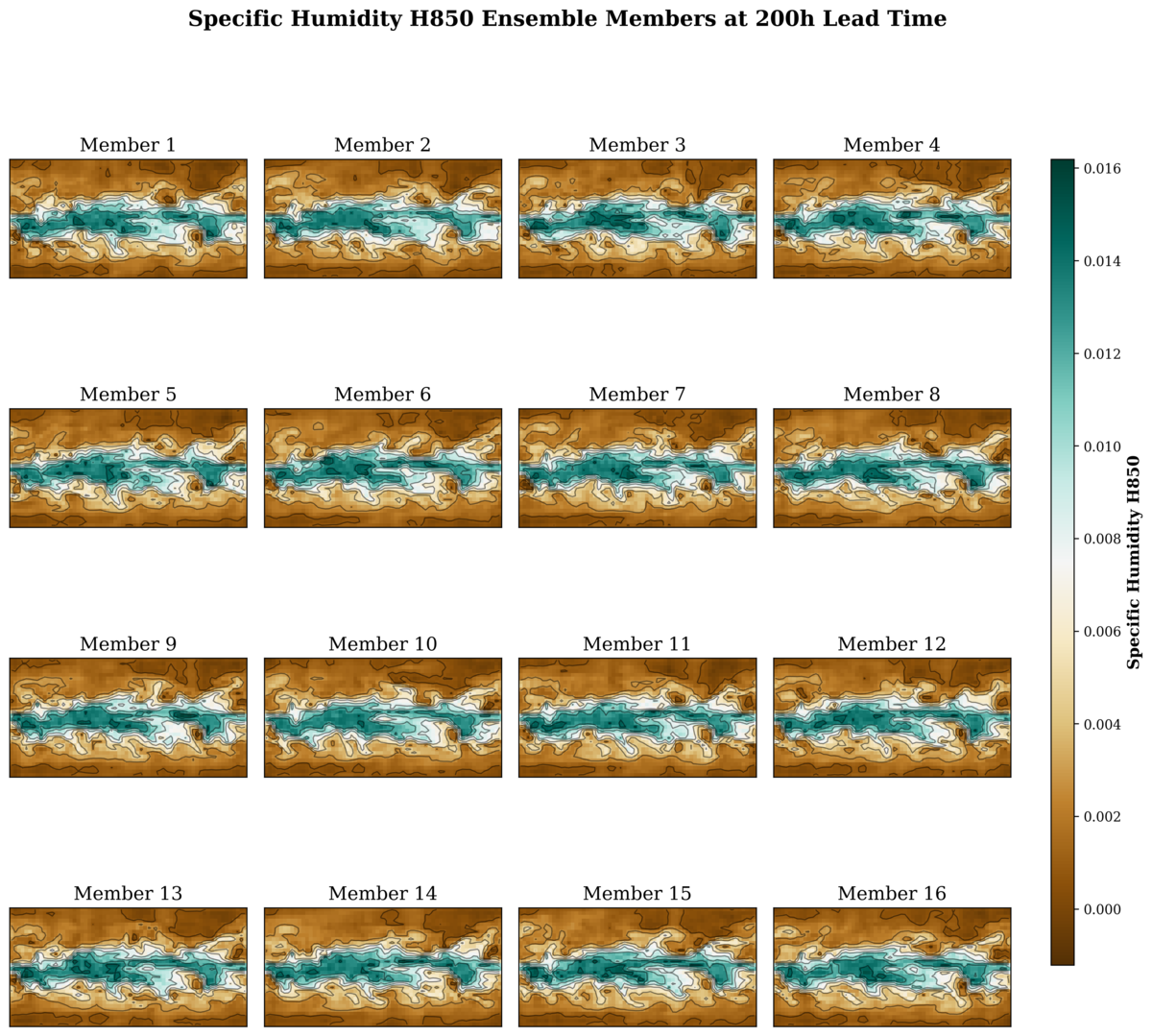}
        \label{fig:members_humid}
    \end{subfigure}
    \hfill
    \begin{subfigure}[b]{0.48\textwidth}
        \centering
        \includegraphics[width=\textwidth]{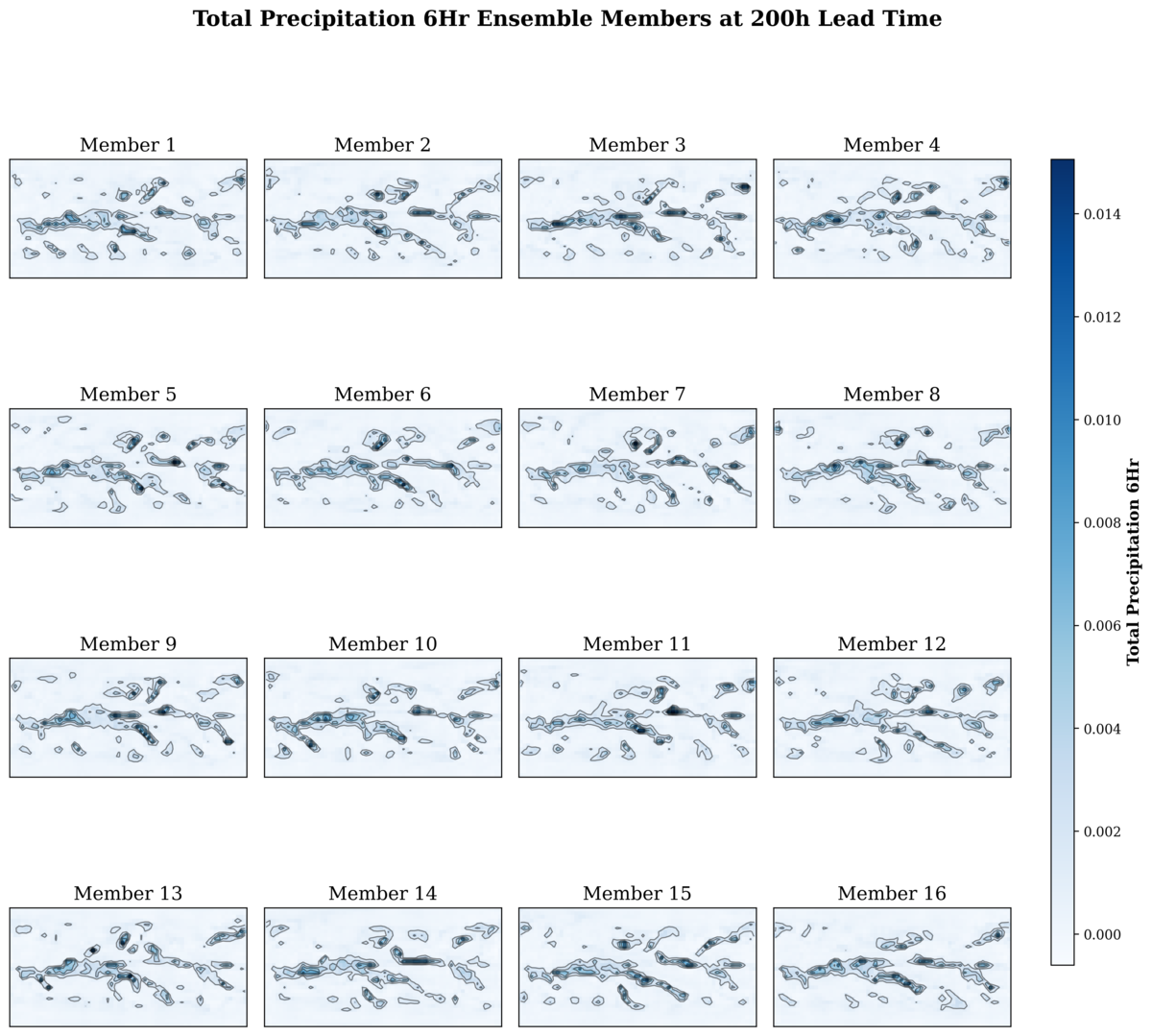}
        \label{fig:members_rain}
    \end{subfigure}
    \caption{Visualization of 16 randomly selected ensemble members at 200-hour lead time for: (a) 2-meter temperature, (b) 850 hPa geopotential height, (c) 850 hPa specific humidity, and (d) 6-hour accumulated precipitation.}
    \label{fig:ensemble_members_200h}
\end{figure}

We present quantitative results for six variables: 2M Temperature, Geopotential H850, Specific Humidity H850, Total Precipitation, Wind X 10M, and Wind Y 10M. This specific subset was chosen as it best highlights the strengths and limitations of our approach. Analyzing the results from Table \ref{tab:2m_temperature_metrics_guidance}, we observe that the ensemble mean RMSE significantly outperforms the deterministic RMSE for the 2M Temperature variable. This finding is further supported by Figure \ref{fig:rmse_comparison}, which illustrates the magnitude of differences between the deterministic trajectories and the ensemble. While this analysis focuses on a single variable, an examination of additional variables, as presented in the tables in \ref{appendix:additional_tables}, reveals a significant improvement in specific humidity. For other variables, such as geopotential, precipitation, and wind speed, the improvements are less pronounced but still evident. We find that most variables benefit from a guidance term of 0.5 with no iterative updates, whereas variables like geopotential achieve better performance with a higher guidance term of 0.7 and reduced exploration.

Examining the probabilistic metrics in Table \ref{tab:2m_temperature_metrics_guidance}, we observe that the estimates for Energy and CRPS are reasonable. Specifically, these values fall below the mean RMSE, indicating that the ensemble members are sufficiently diverse to capture variations. This observation is further reinforced by the spread correlation, which exhibits a relatively low value, confirming the ensemble's ability to balance accuracy and diversity. These same results can be found within \ref{appendix:additional_tables}.

A qualitative analysis of the results provides deeper insight into the method's behavior. First, examining the deterministic predictions in Figure \ref{fig:deterministic_variables}, we identify two key issues: out-of-bounds values and degenerate long-term trajectories. Specifically, for specific humidity, the method produces negative values over extended time horizons, which fall outside the valid domain. A similar issue arises in the 2M temperature plot, where temperatures near the equator deviate beyond plausible ranges. Comparing the errors observed in Figure \ref{fig:deterministic_variables} with the error plots generated by the ensemble in Figure \ref{fig:comparison_variables}, we observe significantly lower magnitudes. Additionally, the ensemble-based predictions remain within physically plausible bounds. However, one limitation of the ensemble approach is oversmoothing when using the mean of the distribution as the state estimate. This effect is particularly evident in the precipitation and specific humidity variables in Figure \ref{fig:comparison_variables}, where much of the finer detail is lost, leading to an overly smooth estimate.  Figure~\ref{fig:ensemble_members_200h} further illustrates this by displaying 16 randomly selected ensemble members at the 200-hour lead time. As shown, individual members retain sharper features, but their trajectories diverge significantly. As a result, the ensemble mean fails to capture fine-scale structures due to slight overdispersion. While overdispersion is not inherently problematic, it is worth noting in this context. Nonetheless, some finer-scale features, such as the recurring daily drop at the southwestern tip of Africa and extreme heat across Australia remain present, clearly depicted within Figure \ref{fig:comparison_variables}.

\begin{figure}[h!]
    \centering
    \begin{subfigure}[b]{0.48\textwidth}
        \centering
        \includegraphics[width=\textwidth]{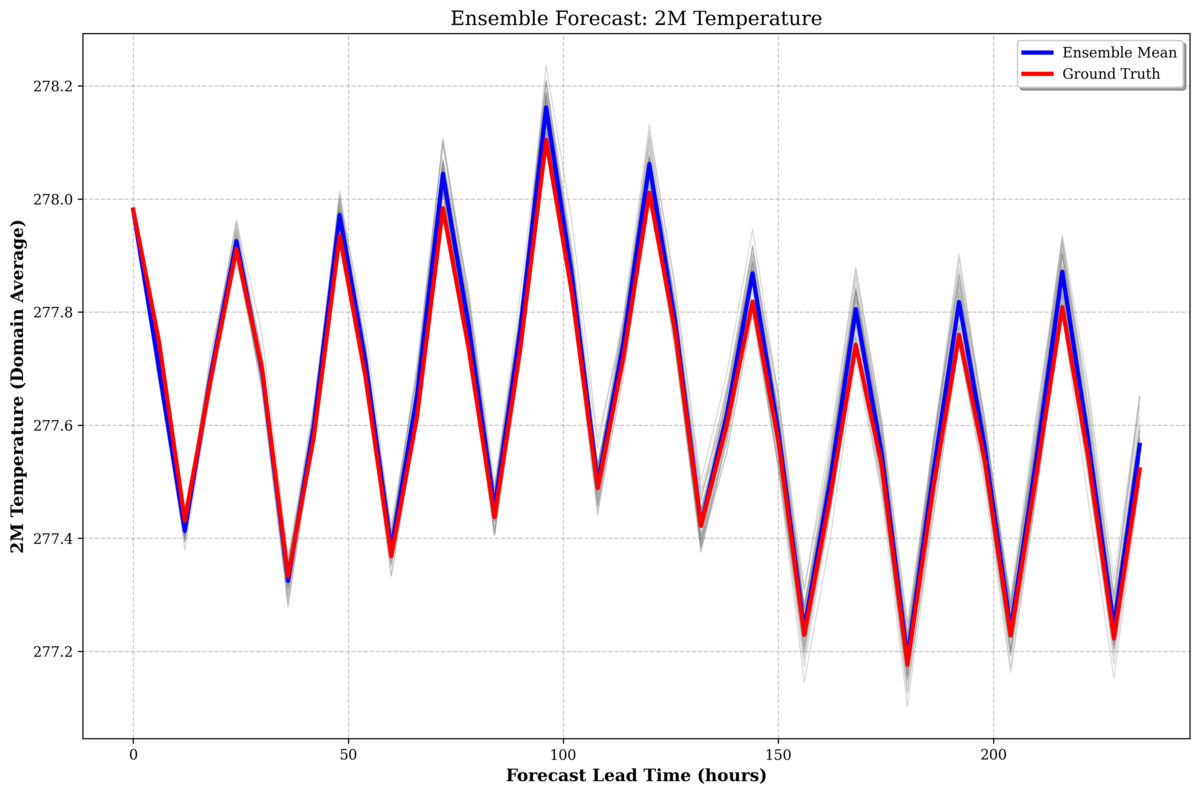}
        \label{fig:energy_temp}
    \end{subfigure}
    \hfill
    \begin{subfigure}[b]{0.48\textwidth}
        \centering
        \includegraphics[width=\textwidth]{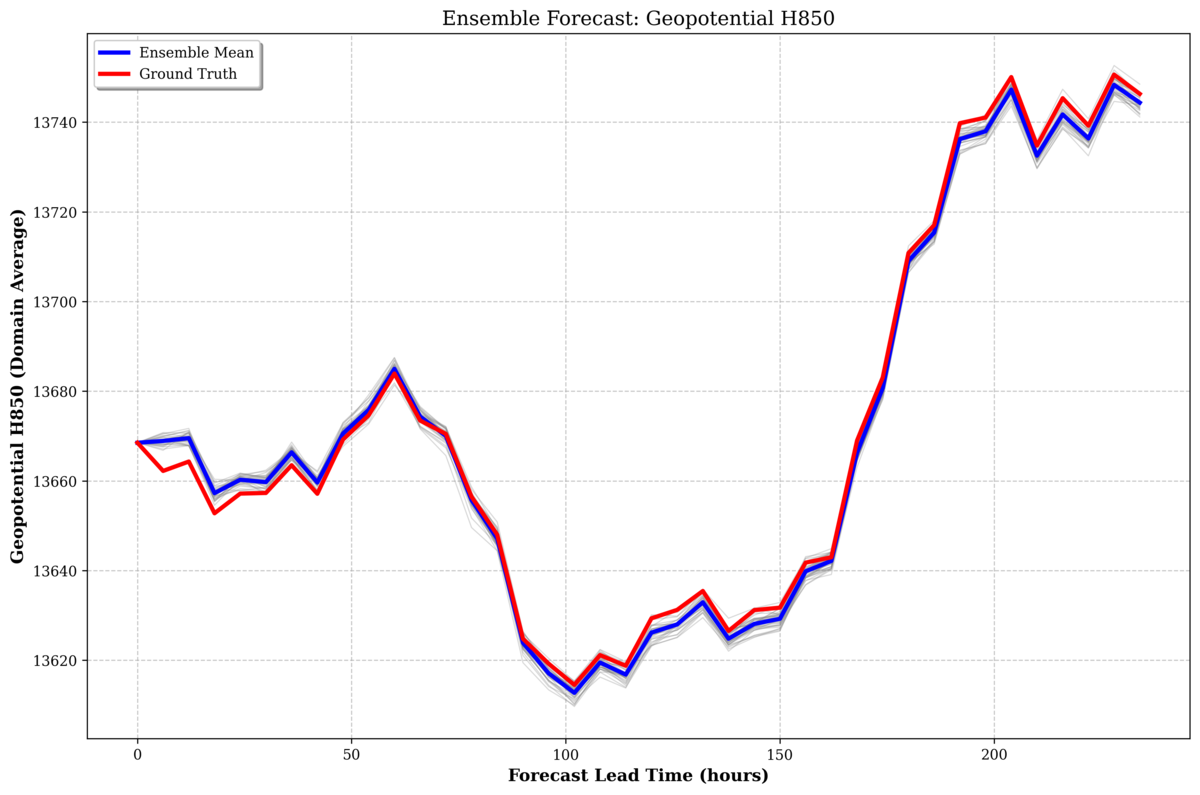}
        \label{fig:energy_geo}
    \end{subfigure}
    
    \vspace{0.5cm}
    
    \begin{subfigure}[b]{0.48\textwidth}
        \centering
        \includegraphics[width=\textwidth]{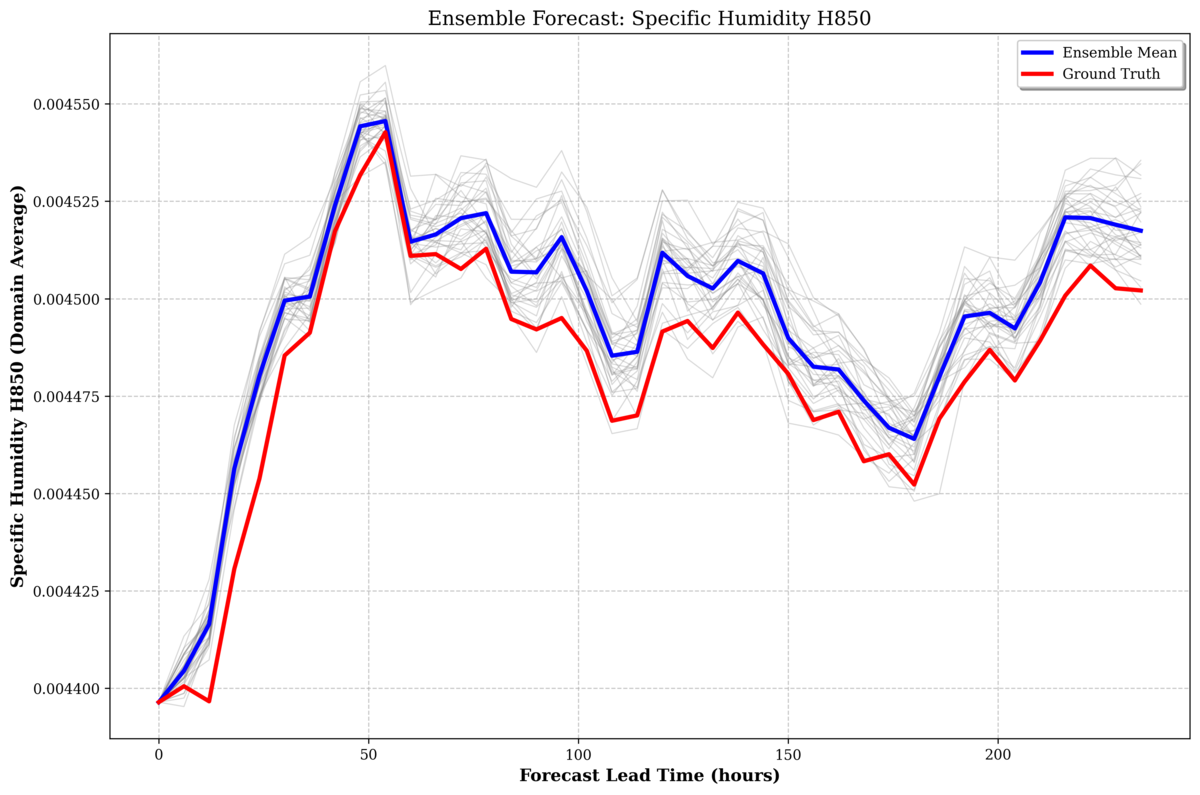}
        \label{fig:energy_humid}
    \end{subfigure}
    \hfill
    \begin{subfigure}[b]{0.48\textwidth}
        \centering
        \includegraphics[width=\textwidth]{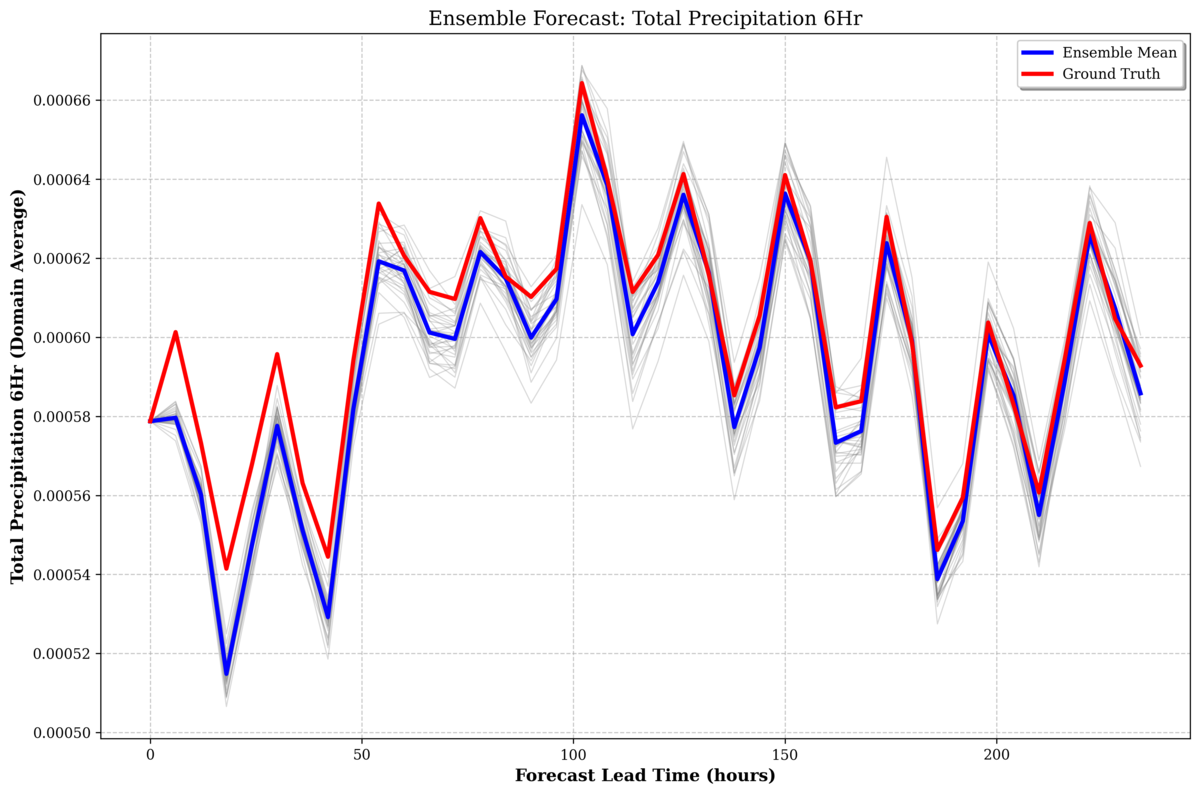}
        \label{fig:energy_rain}
    \end{subfigure}
    \caption{Domain-averaged values of ensemble forecast variables: individual ensemble members (gray lines), ensemble mean prediction (blue line), and ground truth observation (red line) for: (a) 2-meter temperature, (b) 850 hPa geopotential height, (c) 850 hPa specific humidity, and (d) 6-hour accumulated precipitation.}
    \label{fig:domain_averages}
\end{figure}

Comparing the deterministic predictions with each ensemble member in Figure \ref{fig:rmse_comparison}, we observe a slight discrepancy\textendash the deterministic-only method should be contained within the ensemble distribution. However, we find that the deterministic predictions perform significantly worse when evaluated on certain variables, such as specific humidity and temperature. From our previous observations, we noted that the deterministic method produces out-of-domain values. Based on this, we state that an additional benefit of using a diffusion model as a perturbation method is that it also acts as a predictor-corrector, projecting predictions onto a manifold learned from the data distribution. Additionally, we demonstrate the conservation of energy in Figure \ref{fig:domain_averages}. This figure shows that each distribution encompasses the ground truth energy, indicating that while our method may introduce oversmoothing and error accumulation, it effectively preserves the system's energy characteristics.

\section{Discussion}
We find that our approach produces both superior point estimates and well-calibrated uncertainty. Specifically, our method achieves point estimates with a RMSE that is up to 50\% lower on the high end and approximately 10\% lower on the low end, relative to deterministic method, after standardizing to account for the domain's distribution of values. Additionally, our evaluation using the Energy Score and CRPS demonstrates that our method generates uncertainty estimates (spreads) that accurately reflect the broad variability in trajectory space, while remaining concentrated around a central, physically plausible direction.

By incorporating a DDPM-based perturbation mechanism into a strong/weak time advancement scheme, we achieve an effective, parallelizable, and, most importantly, intuitive ensemble forecasting approach. Qualitatively, we observe that the resulting distribution captures accumulated error well enough to provide a rough estimate of regions with higher uncertainty. While our experiments produce smooth long-term forecasts, these could be further improved by integrating a stronger deterministic model. Notably, the diffusion model itself does not need to be highly accurate, as its role is straightforward yet non-trivial. An additional feature of this framework is that it enables transforming any machine learning-based deterministic integrator into a stochastic system that provides both pointwise forecasts and uncertainty estimates over the trajectory distribution. We demonstrate that even our weak deterministic predictor, when combined with the diffusion-based perturbation model, conserves energy and adheres to valid domain constraints—unlike the purely deterministic model alone. Our extension acts as a predictor-corrector, enabling nonlinear projection of predictions onto a learned manifold derived from the training distribution. This improves upon prior methods that approximate such projections using spatial pointwise clipping or dimensionality reduction techniques. We consider this approach more robust, as it leverages the parameterized latent distributions to capture both high- and low-level structures. An added benefit of these distributions is their ability to represent multi-resolution patterns that many modern methods fail to capture.

\noindent\textbf{Limitations.} While the proposed framework is evaluated using a weak deterministic integrator, its results are not directly comparable to current state-of-the-art methods. Due to limited computational resources, we do not make state-of-the-art claims and therefore omit comparisons against benchmarks such as WeatherBench. Our claim of generalizability is based on architectural design and has not been empirically validated. We also acknowledge the computational demands and slower inference speeds inherent to diffusion-based models. Although we mitigate this by employing the DPM++ solver, which improves inference times by approximately two orders of magnitude, this comes at the cost of losing fine-grained information due to integrating over latent variables in the Markov chain. Lastly, while the framework is designed to be generalizable, real-world systems often involve different data transformations and subsets. For example, operational ERA5 includes fewer variables than the ERA5 Reanalysis dataset. To ensure broader applicability, a family of models would likely need to be trained across various subsets, and researchers may need to apply custom post-processing steps to adapt the model outputs to their specific operational settings.

\section{Conclusion and Future Work}

In this paper, we present an approach for performing condition perturbation using a DDPM. This method enables any deterministic, machine learning-based predictor to generate stochastic forecasts. Leveraging the DDPM formulation, we introduce two mechanisms to balance exploration and exploitation: a guidance term and iterative sampling (or walks). We find that this method produces forecasts that outperform the deterministic baseline and has the added benefit of acting as a predictor-corrector system. Since the DDPM generates perturbations inherently, the method is highly parallelizable and, once trained, requires only compute resources—unlike traditional data assimilation techniques that depend on additional observational data.

A natural next step is to evaluate this method in conjunction with a strong predictor, such as GraphCast or FourCastNet. This would enable benchmarking against state-of-the-art methods using standardized benchmarks such as WeatherBench. Given the promising results, we also propose an extension to improve the selection of the most accurate forecast trajectory. While our current approach relies on the pointwise mean, we hypothesize that selecting the trajectory with the highest associated probability may yield more robust forecasts. To support this, we propose learning a scalar quantity that quantifies a state’s relative proximity to the ground truth, modeled as a state-value function parameterized by a neural network. The method would evaluate each ensemble member independently and output a distribution over relative uncertainties. To avoid collapse of this distribution, we introduce a regularization term based on the Kullback–Leibler (KL) divergence with respect to a prior, such as a standard normal distribution. Additionally, we propose a slight modification to this prior formulation. For highly probable states, exploration can be enhanced using techniques such as Monte Carlo Tree Search (MCTS) or beam search, allowing us to refine our sampling strategy toward multiple likely trajectories instead of covering the entire distribution, which may include divergent paths. Rather than generating $N$ independent trajectories from $N$ copies of the initial condition, this approach selectively expands the most probable trajectories while pruning low-probability paths, resulting in a tree-like structure for forecast refinement.


 \bibliographystyle{elsarticle-num} 
 \bibliography{ref}


\newpage
\appendix

\section{Code Availability}\label{appendix:repo}
Our repository is available under the Apache License version 2.0 at \url{https://github.com/djm3622/def-era}. Our checkpoints and output (including plots, tables, and animations) are too large to be included in the repository, as the total size is approximately 5 GB for the checkpoints and 39 GB for the output. However, they can be obtained upon request from the corresponding author.

\section{Additional Tables}\label{appendix:additional_tables}

\begin{table}[h]
    \centering
    \setlength{\tabcolsep}{3pt} 
    \scriptsize 
    \begin{tabular}{l S S S S | S S S S }
        \toprule
        & \multicolumn{4}{c|}{\textbf{Lead Time: 60h}} & \multicolumn{4}{c}{\textbf{Lead Time: 120h}} \\
        \cmidrule(lr){2-5} \cmidrule(lr){6-9}
        & \multicolumn{1}{c}{Energy} & \multicolumn{1}{c}{CRPS} & \multicolumn{1}{c}{RMSE} & \multicolumn{1}{c|}{Spread} & \multicolumn{1}{c}{Energy} & \multicolumn{1}{c}{CRPS} & \multicolumn{1}{c}{RMSE} & \multicolumn{1}{c}{Spread} \\
        \midrule
        Deterministic & {-} & {-} & \num{353.98} & {-} & {-} & {-} & \num{661.43} & {-} \\
        Diffusion[0.3, 1] & \num{314.64} & \num{218.08} & \num{579.15} & \num{16293.34} & \num{481.39} & \num{352.08} & \num{778.16} & \num{23082.36} \\
        Diffusion[0.5, 1] & \textbf{\num{209.00}} & \textbf{\num{156.14}} & \num{365.31} & \textbf{\num{10512.64}} & \textbf{\num{415.48}} & \textbf{\num{289.55}} & \num{657.92} & \num{20338.36} \\
        Diffusion[0.7, 1] & \num{258.51} & \num{198.95} & \textbf{\num{322.95}} & \num{11543.85} & \num{439.44} & \num{308.87} & \textbf{\num{594.52}} & \textbf{\num{20007.06}} \\
        Diffusion[1.0, 1] & \num{328.93} & \num{253.57} & \num{342.88} & \num{14804.79} & \num{617.28} & \num{452.28} & \num{652.89} & \num{27822.31} \\
        \midrule
        & \multicolumn{4}{c|}{\textbf{Lead Time: 180h}} & \multicolumn{4}{c}{\textbf{Lead Time: 234h}} \\
        \cmidrule(lr){2-5} \cmidrule(lr){6-9}
        & \multicolumn{1}{c}{Energy} & \multicolumn{1}{c}{CRPS} & \multicolumn{1}{c}{RMSE} & \multicolumn{1}{c|}{Spread} & \multicolumn{1}{c}{Energy} & \multicolumn{1}{c}{CRPS} & \multicolumn{1}{c}{RMSE} & \multicolumn{1}{c}{Spread} \\
        \midrule
        Deterministic & {-} & {-} & \num{733.33} & {-} & {-} & {-} & \num{718.15} & {-} \\
        Diffusion[0.3, 1] & \num{478.23} & \num{349.17} & \num{742.92} & \num{22514.09} & \num{471.72} & \num{330.93} & \num{718.45} & \num{21736.51} \\
        Diffusion[0.5, 1] & \num{487.00} & \textbf{\num{345.84}} & \num{720.82} & \num{22708.24} & \num{443.12} & \num{301.71} & \num{680.74} & \num{20721.48} \\
        Diffusion[0.7, 1] & \textbf{\num{456.94}} & \num{352.36} & \textbf{\num{657.62}} & \textbf{\num{21437.25}} & \textbf{\num{418.87}} & \textbf{\num{294.97}} & \textbf{\num{657.02}} & \textbf{\num{20171.60}} \\
        Diffusion[1.0, 1] & \num{681.70} & \num{511.53} & \num{739.08} & \num{30627.15} & \num{681.87} & \num{478.10} & \num{758.08} & \num{30736.99} \\
        \bottomrule
    \end{tabular}
    \caption{Geopotential H850 forecast metrics across different lead times.}
    \label{tab:geopotential_h850_metrics}
\end{table}
\begin{table}[h]
    \centering
    \setlength{\tabcolsep}{3pt} 
    \scriptsize 
    \begin{tabular}{l S S S S | S S S S }
        \toprule
        & \multicolumn{4}{c|}{\textbf{Lead Time: 60h}} & \multicolumn{4}{c}{\textbf{Lead Time: 120h}} \\
        \cmidrule(lr){2-5} \cmidrule(lr){6-9}
        & \multicolumn{1}{c}{Energy} & \multicolumn{1}{c}{CRPS} & \multicolumn{1}{c}{RMSE} & \multicolumn{1}{c|}{Spread} & \multicolumn{1}{c}{Energy} & \multicolumn{1}{c}{CRPS} & \multicolumn{1}{c}{RMSE} & \multicolumn{1}{c}{Spread} \\
        \midrule
        Deterministic & {-} & {-} & \num{0.00152} & {-} & {-} & {-} & \num{0.00239} & {-} \\
        Diffusion[0.3, 1] & \num{0.00079} & \num{0.00055} & \num{0.00153} & \num{0.04350} & \num{0.00093} & \num{0.00064} & \num{0.00167} & \num{0.04830} \\
        Diffusion[0.5, 1] & \textbf{\num{0.00076}} & \textbf{\num{0.00054}} & \textbf{\num{0.00122}} & \textbf{\num{0.03738}} & \textbf{\num{0.00081}} & \textbf{\num{0.00056}} & \textbf{\num{0.00136}} & \textbf{\num{0.04254}} \\
        Diffusion[0.7, 1] & \num{0.00099} & \num{0.00069} & \num{0.00123} & \num{0.04462} & \num{0.00110} & \num{0.00078} & \num{0.00147} & \num{0.05148} \\
        Diffusion[1.0, 1] & \num{0.00153} & \num{0.00104} & \num{0.00161} & \num{0.06889} & \num{0.00229} & \num{0.00153} & \num{0.00244} & \num{0.10296} \\
        \midrule
        & \multicolumn{4}{c|}{\textbf{Lead Time: 180h}} & \multicolumn{4}{c}{\textbf{Lead Time: 234h}} \\
        \cmidrule(lr){2-5} \cmidrule(lr){6-9}
        & \multicolumn{1}{c}{Energy} & \multicolumn{1}{c}{CRPS} & \multicolumn{1}{c}{RMSE} & \multicolumn{1}{c|}{Spread} & \multicolumn{1}{c}{Energy} & \multicolumn{1}{c}{CRPS} & \multicolumn{1}{c}{RMSE} & \multicolumn{1}{c}{Spread} \\
        \midrule
        Deterministic & {-} & {-} & \num{0.00361} & {-} & {-} & {-} & \num{0.00410} & {-} \\
        Diffusion[0.3, 1] & \num{0.00089} & \num{0.00065} & \num{0.00164} & \num{0.04824} & \num{0.00097} & \num{0.00066} & \num{0.00172} & \num{0.05132} \\
        Diffusion[0.5, 1] & \textbf{\num{0.00084}} & \textbf{\num{0.00061}} & \textbf{\num{0.00144}} & \textbf{\num{0.04395}} & \textbf{\num{0.00082}} & \textbf{\num{0.00060}} & \textbf{\num{0.00144}} & \textbf{\num{0.04438}} \\
        Diffusion[0.7, 1] & \num{0.00103} & \num{0.00076} & \num{0.00153} & \num{0.05020} & \num{0.00108} & \num{0.00077} & \num{0.00166} & \num{0.05442} \\
        Diffusion[1.0, 1] & \num{0.00297} & \num{0.00191} & \num{0.00322} & \num{0.13318} & \num{0.00341} & \num{0.00224} & \num{0.00381} & \num{0.15232} \\
        \bottomrule
    \end{tabular}
    \caption{Specific Humidity H850 forecast metrics across different lead times.}
    \label{tab:specific_humidity_h850_metrics}
\end{table}
\begin{table}[h]
    \centering
    \setlength{\tabcolsep}{3pt} 
    \scriptsize 
    \begin{tabular}{l S S S S | S S S S }
        \toprule
        & \multicolumn{4}{c|}{\textbf{Lead Time: 60h}} & \multicolumn{4}{c}{\textbf{Lead Time: 120h}} \\
        \cmidrule(lr){2-5} \cmidrule(lr){6-9}
        & \multicolumn{1}{c}{Energy} & \multicolumn{1}{c}{CRPS} & \multicolumn{1}{c}{RMSE} & \multicolumn{1}{c|}{Spread} & \multicolumn{1}{c}{Energy} & \multicolumn{1}{c}{CRPS} & \multicolumn{1}{c}{RMSE} & \multicolumn{1}{c}{Spread} \\
        \midrule
        Deterministic & {-} & {-} & \num{0.00140} & {-} & {-} & {-} & \num{0.00181} & {-} \\
        Diffusion[0.3, 1] & \num{0.00065} & \textbf{\num{0.00029}} & \num{0.00134} & \num{0.04487} & \textbf{\num{0.00072}} & \textbf{\num{0.00033}} & \num{0.00147} & \num{0.04999} \\
        Diffusion[0.5, 1] & \textbf{\num{0.00064}} & \num{0.00030} & \textbf{\num{0.00118}} & \textbf{\num{0.03625}} & \num{0.00074} & \num{0.00038} & \textbf{\num{0.00140}} & \textbf{\num{0.04475}} \\
        Diffusion[0.7, 1] & \num{0.00093} & \num{0.00041} & \num{0.00122} & \num{0.04342} & \num{0.00118} & \num{0.00054} & \num{0.00165} & \num{0.05853} \\
        Diffusion[1.0, 1] & \num{0.00139} & \num{0.00058} & \num{0.00147} & \num{0.06283} & \num{0.00166} & \num{0.00080} & \num{0.00181} & \num{0.07541} \\
        \midrule
        & \multicolumn{4}{c|}{\textbf{Lead Time: 180h}} & \multicolumn{4}{c}{\textbf{Lead Time: 234h}} \\
        \cmidrule(lr){2-5} \cmidrule(lr){6-9}
        & \multicolumn{1}{c}{Energy} & \multicolumn{1}{c}{CRPS} & \multicolumn{1}{c}{RMSE} & \multicolumn{1}{c|}{Spread} & \multicolumn{1}{c}{Energy} & \multicolumn{1}{c}{CRPS} & \multicolumn{1}{c}{RMSE} & \multicolumn{1}{c}{Spread} \\
        \midrule
        Deterministic & {-} & {-} & \num{0.00183} & {-} & {-} & {-} & \num{0.00182} & {-} \\
        Diffusion[0.3, 1] & \textbf{\num{0.00072}} & \textbf{\num{0.00031}} & \textbf{\num{0.00141}} & \textbf{\num{0.05018}} & \textbf{\num{0.00074}} & \textbf{\num{0.00034}} & \textbf{\num{0.00143}} & \num{0.05120} \\
        Diffusion[0.5, 1] & \num{0.00083} & \num{0.00039} & \num{0.00148} & \num{0.05121} & \num{0.00082} & \num{0.00037} & \num{0.00147} & \textbf{\num{0.05106}} \\
        Diffusion[0.7, 1] & \num{0.00104} & \num{0.00047} & \num{0.00160} & \num{0.05751} & \num{0.00094} & \num{0.00043} & \num{0.00155} & \num{0.05612} \\
        Diffusion[1.0, 1] & \num{0.00161} & \num{0.00079} & \num{0.00182} & \num{0.07453} & \num{0.00153} & \num{0.00078} & \num{0.00179} & \num{0.07144} \\
        \bottomrule
    \end{tabular}
    \caption{Total Precipitation 6Hr forecast metrics across different lead times.}
    \label{tab:total_precipitation_6hr_metrics}
\end{table}
\begin{table}[h]
    \centering
    \setlength{\tabcolsep}{3pt} 
    \scriptsize 
    \begin{tabular}{l S S S S | S S S S }
        \toprule
        & \multicolumn{4}{c|}{\textbf{Lead Time: 60h}} & \multicolumn{4}{c}{\textbf{Lead Time: 120h}} \\
        \cmidrule(lr){2-5} \cmidrule(lr){6-9}
        & \multicolumn{1}{c}{Energy} & \multicolumn{1}{c}{CRPS} & \multicolumn{1}{c}{RMSE} & \multicolumn{1}{c|}{Spread} & \multicolumn{1}{c}{Energy} & \multicolumn{1}{c}{CRPS} & \multicolumn{1}{c}{RMSE} & \multicolumn{1}{c}{Spread} \\
        \midrule
        Deterministic & {-} & {-} & \num{2.06} & {-} & {-} & {-} & \num{3.85} & {-} \\
        Diffusion[0.3, 1] & \num{1.44} & \num{0.97} & \num{3.04} & \num{88.47} & \textbf{\num{1.94}} & \textbf{\num{1.36}} & \num{3.62} & \textbf{\num{106.00}} \\
        Diffusion[0.5, 1] & \textbf{\num{1.07}} & \textbf{\num{0.82}} & \num{2.11} & \textbf{\num{61.77}} & \num{2.11} & \num{1.52} & \textbf{\num{3.50}} & \num{109.27} \\
        Diffusion[0.7, 1] & \num{1.44} & \num{1.09} & \textbf{\num{1.93}} & \num{66.29} & \num{2.58} & \num{1.84} & \num{3.54} & \num{121.30} \\
        Diffusion[1.0, 1] & \num{1.98} & \num{1.47} & \num{2.10} & \num{89.39} & \num{3.57} & \num{2.54} & \num{3.83} & \num{161.62} \\
        \midrule
        & \multicolumn{4}{c|}{\textbf{Lead Time: 180h}} & \multicolumn{4}{c}{\textbf{Lead Time: 234h}} \\
        \cmidrule(lr){2-5} \cmidrule(lr){6-9}
        & \multicolumn{1}{c}{Energy} & \multicolumn{1}{c}{CRPS} & \multicolumn{1}{c}{RMSE} & \multicolumn{1}{c|}{Spread} & \multicolumn{1}{c}{Energy} & \multicolumn{1}{c}{CRPS} & \multicolumn{1}{c}{RMSE} & \multicolumn{1}{c}{Spread} \\
        \midrule
        Deterministic & {-} & {-} & \num{3.77} & {-} & {-} & {-} & \num{4.22} & {-} \\
        Diffusion[0.3, 1] & \textbf{\num{1.76}} & \textbf{\num{1.31}} & \num{3.29} & \textbf{\num{93.53}} & \textbf{\num{2.13}} & \textbf{\num{1.48}} & \num{3.65} & \textbf{\num{106.98}} \\
        Diffusion[0.5, 1] & \num{1.90} & \num{1.44} & \num{3.26} & \num{98.74} & \num{2.16} & \num{1.50} & \textbf{\num{3.60}} & \num{110.20} \\
        Diffusion[0.7, 1] & \num{1.96} & \num{1.52} & \textbf{\num{3.16}} & \num{102.63} & \num{2.33} & \num{1.72} & \num{3.76} & \num{119.42} \\
        Diffusion[1.0, 1] & \num{3.54} & \num{2.66} & \num{3.93} & \num{159.99} & \num{3.86} & \num{2.94} & \num{4.39} & \num{173.94} \\
        \bottomrule
    \end{tabular}
    \caption{Wind X 10M forecast metrics across different lead times.}
    \label{tab:wind_x_10m_metrics}
\end{table}
\begin{table}[h]
    \centering
    \setlength{\tabcolsep}{3pt} 
    \scriptsize 
    \begin{tabular}{l S S S S | S S S S }
        \toprule
        & \multicolumn{4}{c|}{\textbf{Lead Time: 60h}} & \multicolumn{4}{c}{\textbf{Lead Time: 120h}} \\
        \cmidrule(lr){2-5} \cmidrule(lr){6-9}
        & \multicolumn{1}{c}{Energy} & \multicolumn{1}{c}{CRPS} & \multicolumn{1}{c}{RMSE} & \multicolumn{1}{c|}{Spread} & \multicolumn{1}{c}{Energy} & \multicolumn{1}{c}{CRPS} & \multicolumn{1}{c}{RMSE} & \multicolumn{1}{c}{Spread} \\
        \midrule
        Deterministic & {-} & {-} & \num{2.33} & {-} & {-} & {-} & \num{4.06} & {-} \\
        Diffusion[0.3, 1] & \num{1.42} & \num{0.96} & \num{3.08} & \num{86.65} & \num{2.10} & \textbf{\num{1.54}} & \num{3.88} & \num{108.05} \\
        Diffusion[0.5, 1] & \textbf{\num{1.19}} & \textbf{\num{0.91}} & \num{2.32} & \textbf{\num{67.35}} & \textbf{\num{2.04}} & \num{1.60} & \num{3.68} & \textbf{\num{107.12}} \\
        Diffusion[0.7, 1] & \num{1.55} & \num{1.16} & \textbf{\num{2.08}} & \num{70.19} & \num{2.45} & \num{1.81} & \textbf{\num{3.60}} & \num{116.99} \\
        Diffusion[1.0, 1] & \num{2.24} & \num{1.62} & \num{2.37} & \num{101.11} & \num{3.96} & \num{2.80} & \num{4.24} & \num{179.07} \\
        \midrule
        & \multicolumn{4}{c|}{\textbf{Lead Time: 180h}} & \multicolumn{4}{c}{\textbf{Lead Time: 234h}} \\
        \cmidrule(lr){2-5} \cmidrule(lr){6-9}
        & \multicolumn{1}{c}{Energy} & \multicolumn{1}{c}{CRPS} & \multicolumn{1}{c}{RMSE} & \multicolumn{1}{c|}{Spread} & \multicolumn{1}{c}{Energy} & \multicolumn{1}{c}{CRPS} & \multicolumn{1}{c}{RMSE} & \multicolumn{1}{c}{Spread} \\
        \midrule
        Deterministic & {-} & {-} & \num{4.27} & {-} & {-} & {-} & \num{4.58} & {-} \\
        Diffusion[0.3, 1] & \textbf{\num{2.09}} & \textbf{\num{1.47}} & \textbf{\num{3.75}} & \textbf{\num{110.02}} & \textbf{\num{2.29}} & \textbf{\num{1.54}} & \textbf{\num{4.06}} & \textbf{\num{119.75}} \\
        Diffusion[0.5, 1] & \num{2.46} & \num{1.72} & \num{3.96} & \num{124.53} & \num{2.41} & \num{1.67} & \num{4.07} & \num{125.59} \\
        Diffusion[0.7, 1] & \num{2.67} & \num{1.93} & \num{4.05} & \num{129.21} & \num{2.45} & \num{1.81} & \num{4.11} & \num{123.72} \\
        Diffusion[1.0, 1] & \num{4.25} & \num{3.12} & \num{4.72} & \num{192.14} & \num{4.27} & \num{3.21} & \num{4.90} & \num{192.13} \\
        \bottomrule
    \end{tabular}
    \caption{Wind Y 10M forecast metrics across different lead times.}
    \label{tab:wind_y_10m_metrics}
\end{table}

\begin{table}[h]
    \centering
    \setlength{\tabcolsep}{3pt} 
    \scriptsize 
    \begin{tabular}{l S S S S | S S S S }
        \toprule
        & \multicolumn{4}{c|}{\textbf{Lead Time: 60h}} & \multicolumn{4}{c}{\textbf{Lead Time: 120h}} \\
        \cmidrule(lr){2-5} \cmidrule(lr){6-9}
        & \multicolumn{1}{c}{Energy} & \multicolumn{1}{c}{CRPS} & \multicolumn{1}{c}{RMSE} & \multicolumn{1}{c|}{Spread} & \multicolumn{1}{c}{Energy} & \multicolumn{1}{c}{CRPS} & \multicolumn{1}{c}{RMSE} & \multicolumn{1}{c}{Spread} \\
        \midrule
        Deterministic & {-} & {-} & \num{3.55} & {-} & {-} & {-} & \num{5.28} & {-} \\
        Diffusion[0.5, 1] & \textbf{\num{1.89}} & \textbf{\num{1.52}} & \textbf{\num{2.88}} & \textbf{\num{86.88}} & \textbf{\num{1.62}} & \textbf{\num{1.33}} & \textbf{\num{2.99}} & \textbf{\num{84.78}} \\
        Diffusion[0.5, 3] & \num{2.00} & \num{1.59} & \num{3.23} & \num{95.13} & \num{1.92} & \num{1.41} & \num{3.33} & \num{93.91} \\
        Diffusion[0.5, 7] & \num{2.35} & \num{1.86} & \num{3.76} & \num{106.30} & \num{2.35} & \num{1.68} & \num{3.78} & \num{115.11} \\
        \midrule
        & \multicolumn{4}{c|}{\textbf{Lead Time: 180h}} & \multicolumn{4}{c}{\textbf{Lead Time: 234h}} \\
        \cmidrule(lr){2-5} \cmidrule(lr){6-9}
        & \multicolumn{1}{c}{Energy} & \multicolumn{1}{c}{CRPS} & \multicolumn{1}{c}{RMSE} & \multicolumn{1}{c|}{Spread} & \multicolumn{1}{c}{Energy} & \multicolumn{1}{c}{CRPS} & \multicolumn{1}{c}{RMSE} & \multicolumn{1}{c}{Spread} \\
        \midrule
        Deterministic & {-} & {-} & \num{7.31} & {-} & {-} & {-} & \num{8.12} & {-} \\
        Diffusion[0.5, 1] & \textbf{\num{2.28}} & \textbf{\num{1.52}} & \textbf{\num{3.78}} & \textbf{\num{114.75}} & \textbf{\num{2.42}} & \textbf{\num{1.62}} & \num{4.03} & \textbf{\num{118.24}} \\
        Diffusion[0.5, 3] & \num{2.60} & \num{1.79} & \num{3.99} & \num{124.17} & \num{2.63} & \num{1.84} & \textbf{\num{4.01}} & \num{122.05} \\
        Diffusion[0.5, 7] & \num{2.73} & \num{2.04} & \num{4.23} & \num{130.69} & \num{2.59} & \num{1.84} & \num{4.07} & \num{122.75} \\
        \bottomrule
    \end{tabular}
    \caption{2M Temperature forecast metrics across different lead times.}
    \label{tab:2m_temperature_metrics}
\end{table}
\begin{table}[h]
    \centering
    \setlength{\tabcolsep}{3pt} 
    \scriptsize 
    \begin{tabular}{l S S S S | S S S S }
        \toprule
        & \multicolumn{4}{c|}{\textbf{Lead Time: 60h}} & \multicolumn{4}{c}{\textbf{Lead Time: 120h}} \\
        \cmidrule(lr){2-5} \cmidrule(lr){6-9}
        & \multicolumn{1}{c}{Energy} & \multicolumn{1}{c}{CRPS} & \multicolumn{1}{c}{RMSE} & \multicolumn{1}{c|}{Spread} & \multicolumn{1}{c}{Energy} & \multicolumn{1}{c}{CRPS} & \multicolumn{1}{c}{RMSE} & \multicolumn{1}{c}{Spread} \\
        \midrule
        Deterministic & {-} & {-} & \num{353.98} & {-} & {-} & {-} & \num{661.43} & {-} \\
        Diffusion[0.5, 1] & \textbf{\num{209.00}} & \textbf{\num{160.64}} & \textbf{\num{365.31}} & \textbf{\num{10512.64}} & \textbf{\num{415.48}} & \textbf{\num{280.70}} & \textbf{\num{657.92}} & \textbf{\num{20338.36}} \\
        Diffusion[0.5, 3] & \num{294.18} & \num{220.47} & \num{481.75} & \num{14571.69} & \num{518.82} & \num{376.91} & \num{745.78} & \num{23917.71} \\
        Diffusion[0.5, 7] & \num{351.64} & \num{261.79} & \num{547.50} & \num{16861.16} & \num{572.41} & \num{420.69} & \num{785.19} & \num{26484.67} \\
        \midrule
        & \multicolumn{4}{c|}{\textbf{Lead Time: 180h}} & \multicolumn{4}{c}{\textbf{Lead Time: 234h}} \\
        \cmidrule(lr){2-5} \cmidrule(lr){6-9}
        & \multicolumn{1}{c}{Energy} & \multicolumn{1}{c}{CRPS} & \multicolumn{1}{c}{RMSE} & \multicolumn{1}{c|}{Spread} & \multicolumn{1}{c}{Energy} & \multicolumn{1}{c}{CRPS} & \multicolumn{1}{c}{RMSE} & \multicolumn{1}{c}{Spread} \\
        \midrule
        Deterministic & {-} & {-} & \num{733.33} & {-} & {-} & {-} & \num{718.15} & {-} \\
        Diffusion[0.5, 1] & \textbf{\num{487.00}} & \textbf{\num{344.54}} & \textbf{\num{720.82}} & \textbf{\num{22708.24}} & \textbf{\num{443.12}} & \textbf{\num{310.13}} & \textbf{\num{680.74}} & \textbf{\num{20721.48}} \\
        Diffusion[0.5, 3] & \num{566.81} & \num{384.14} & \num{760.41} & \num{26024.32} & \num{506.56} & \num{342.19} & \num{700.45} & \num{22271.69} \\
        Diffusion[0.5, 7] & \num{554.41} & \num{393.19} & \num{751.98} & \num{25811.24} & \num{505.04} & \num{361.78} & \num{695.02} & \num{22926.93} \\
        \bottomrule
    \end{tabular}
    \caption{Geopotential H850 forecast metrics across different lead times.}
    \label{tab:geopotential_h850_metrics}
\end{table}
\begin{table}[h]
    \centering
    \setlength{\tabcolsep}{3pt} 
    \scriptsize 
    \begin{tabular}{l S S S S | S S S S }
        \toprule
        & \multicolumn{4}{c|}{\textbf{Lead Time: 60h}} & \multicolumn{4}{c}{\textbf{Lead Time: 120h}} \\
        \cmidrule(lr){2-5} \cmidrule(lr){6-9}
        & \multicolumn{1}{c}{Energy} & \multicolumn{1}{c}{CRPS} & \multicolumn{1}{c}{RMSE} & \multicolumn{1}{c|}{Spread} & \multicolumn{1}{c}{Energy} & \multicolumn{1}{c}{CRPS} & \multicolumn{1}{c}{RMSE} & \multicolumn{1}{c}{Spread} \\
        \midrule
        Deterministic & {-} & {-} & \num{0.00152} & {-} & {-} & {-} & \num{0.00239} & {-} \\
        Diffusion[0.5, 1] & \textbf{\num{0.00076}} & \textbf{\num{0.00054}} & \textbf{\num{0.00122}} & \textbf{\num{0.03738}} & \textbf{\num{0.00081}} & \textbf{\num{0.00056}} & \textbf{\num{0.00136}} & \textbf{\num{0.04254}} \\
        Diffusion[0.5, 3] & \num{0.00085} & \num{0.00059} & \num{0.00135} & \num{0.04230} & \num{0.00088} & \num{0.00060} & \num{0.00146} & \num{0.04579} \\
        Diffusion[0.5, 7] & \num{0.00088} & \num{0.00064} & \num{0.00144} & \num{0.04467} & \num{0.00094} & \num{0.00065} & \num{0.00151} & \num{0.04779} \\
        \midrule
        & \multicolumn{4}{c|}{\textbf{Lead Time: 180h}} & \multicolumn{4}{c}{\textbf{Lead Time: 234h}} \\
        \cmidrule(lr){2-5} \cmidrule(lr){6-9}
        & \multicolumn{1}{c}{Energy} & \multicolumn{1}{c}{CRPS} & \multicolumn{1}{c}{RMSE} & \multicolumn{1}{c|}{Spread} & \multicolumn{1}{c}{Energy} & \multicolumn{1}{c}{CRPS} & \multicolumn{1}{c}{RMSE} & \multicolumn{1}{c}{Spread} \\
        \midrule
        Deterministic & {-} & {-} & \num{0.00361} & {-} & {-} & {-} & \num{0.00410} & {-} \\
        Diffusion[0.5, 1] & \textbf{\num{0.00084}} & \textbf{\num{0.00061}} & \textbf{\num{0.00144}} & \textbf{\num{0.04395}} & \textbf{\num{0.00082}} & \textbf{\num{0.00061}} & \num{0.00144} & \num{0.04438} \\
        Diffusion[0.5, 3] & \num{0.00090} & \num{0.00064} & \num{0.00148} & \num{0.04576} & \num{0.00084} & \num{0.00064} & \textbf{\num{0.00142}} & \textbf{\num{0.04330}} \\
        Diffusion[0.5, 7] & \num{0.00093} & \num{0.00068} & \num{0.00152} & \num{0.04767} & \num{0.00087} & \num{0.00066} & \num{0.00145} & \num{0.04486} \\
        \bottomrule
    \end{tabular}
    \caption{Specific Humidity H850 forecast metrics across different lead times.}
    \label{tab:specific_humidity_h850_metrics}
\end{table}
\begin{table}[h]
    \centering
    \setlength{\tabcolsep}{3pt} 
    \scriptsize 
    \begin{tabular}{l S S S S | S S S S }
        \toprule
        & \multicolumn{4}{c|}{\textbf{Lead Time: 60h}} & \multicolumn{4}{c}{\textbf{Lead Time: 120h}} \\
        \cmidrule(lr){2-5} \cmidrule(lr){6-9}
        & \multicolumn{1}{c}{Energy} & \multicolumn{1}{c}{CRPS} & \multicolumn{1}{c}{RMSE} & \multicolumn{1}{c|}{Spread} & \multicolumn{1}{c}{Energy} & \multicolumn{1}{c}{CRPS} & \multicolumn{1}{c}{RMSE} & \multicolumn{1}{c}{Spread} \\
        \midrule
        Deterministic & {-} & {-} & \num{0.00140} & {-} & {-} & {-} & \num{0.00181} & {-} \\
        Diffusion[0.5, 1] & \textbf{\num{0.00064}} & \textbf{\num{0.00030}} & \textbf{\num{0.00118}} & \textbf{\num{0.03625}} & \num{0.00074} & \textbf{\num{0.00037}} & \textbf{\num{0.00140}} & \textbf{\num{0.04475}} \\
        Diffusion[0.5, 3] & \num{0.00071} & \num{0.00034} & \num{0.00130} & \num{0.04011} & \textbf{\num{0.00073}} & \num{0.00039} & \num{0.00142} & \num{0.04536} \\
        Diffusion[0.5, 7] & \num{0.00072} & \num{0.00037} & \num{0.00137} & \num{0.04292} & \num{0.00083} & \num{0.00043} & \num{0.00152} & \num{0.05058} \\
        \midrule
        & \multicolumn{4}{c|}{\textbf{Lead Time: 180h}} & \multicolumn{4}{c}{\textbf{Lead Time: 234h}} \\
        \cmidrule(lr){2-5} \cmidrule(lr){6-9}
        & \multicolumn{1}{c}{Energy} & \multicolumn{1}{c}{CRPS} & \multicolumn{1}{c}{RMSE} & \multicolumn{1}{c|}{Spread} & \multicolumn{1}{c}{Energy} & \multicolumn{1}{c}{CRPS} & \multicolumn{1}{c}{RMSE} & \multicolumn{1}{c}{Spread} \\
        \midrule
        Deterministic & {-} & {-} & \num{0.00183} & {-} & {-} & {-} & \num{0.00182} & {-} \\
        Diffusion[0.5, 1] & \num{0.00083} & \num{0.00039} & \num{0.00148} & \num{0.05121} & \textbf{\num{0.00082}} & \textbf{\num{0.00037}} & \textbf{\num{0.00147}} & \num{0.05106} \\
        Diffusion[0.5, 3] & \num{0.00081} & \textbf{\num{0.00038}} & \textbf{\num{0.00145}} & \textbf{\num{0.04903}} & \num{0.00083} & \num{0.00040} & \num{0.00147} & \textbf{\num{0.05033}} \\
        Diffusion[0.5, 7] & \textbf{\num{0.00080}} & \num{0.00040} & \num{0.00146} & \num{0.04994} & \num{0.00086} & \num{0.00042} & \num{0.00148} & \num{0.05191} \\
        \bottomrule
    \end{tabular}
    \caption{Total Precipitation 6Hr forecast metrics across different lead times.}
    \label{tab:total_precipitation_6hr_metrics}
\end{table}
\begin{table}[h]
    \centering
    \setlength{\tabcolsep}{3pt} 
    \scriptsize 
    \begin{tabular}{l S S S S | S S S S }
        \toprule
        & \multicolumn{4}{c|}{\textbf{Lead Time: 60h}} & \multicolumn{4}{c}{\textbf{Lead Time: 120h}} \\
        \cmidrule(lr){2-5} \cmidrule(lr){6-9}
        & \multicolumn{1}{c}{Energy} & \multicolumn{1}{c}{CRPS} & \multicolumn{1}{c}{RMSE} & \multicolumn{1}{c|}{Spread} & \multicolumn{1}{c}{Energy} & \multicolumn{1}{c}{CRPS} & \multicolumn{1}{c}{RMSE} & \multicolumn{1}{c}{Spread} \\
        \midrule
        Deterministic & {-} & {-} & \num{2.06} & {-} & {-} & {-} & \num{3.85} & {-} \\
        Diffusion[0.5, 1] & \textbf{\num{1.07}} & \textbf{\num{0.84}} & \textbf{\num{2.11}} & \textbf{\num{61.77}} & \textbf{\num{2.11}} & \textbf{\num{1.52}} & \textbf{\num{3.50}} & \textbf{\num{109.27}} \\
        Diffusion[0.5, 3] & \num{1.52} & \num{1.12} & \num{2.70} & \num{79.40} & \num{2.32} & \num{1.69} & \num{3.67} & \num{115.27} \\
        Diffusion[0.5, 7] & \num{1.76} & \num{1.28} & \num{2.97} & \num{88.44} & \num{2.62} & \num{1.87} & \num{3.88} & \num{128.01} \\
        \midrule
        & \multicolumn{4}{c|}{\textbf{Lead Time: 180h}} & \multicolumn{4}{c}{\textbf{Lead Time: 234h}} \\
        \cmidrule(lr){2-5} \cmidrule(lr){6-9}
        & \multicolumn{1}{c}{Energy} & \multicolumn{1}{c}{CRPS} & \multicolumn{1}{c}{RMSE} & \multicolumn{1}{c|}{Spread} & \multicolumn{1}{c}{Energy} & \multicolumn{1}{c}{CRPS} & \multicolumn{1}{c}{RMSE} & \multicolumn{1}{c}{Spread} \\
        \midrule
        Deterministic & {-} & {-} & \num{3.77} & {-} & {-} & {-} & \num{4.22} & {-} \\
        Diffusion[0.5, 1] & \textbf{\num{1.90}} & \textbf{\num{1.41}} & \textbf{\num{3.26}} & \textbf{\num{98.74}} & \textbf{\num{2.16}} & \textbf{\num{1.51}} & \textbf{\num{3.60}} & \textbf{\num{110.20}} \\
        Diffusion[0.5, 3] & \num{2.19} & \num{1.57} & \num{3.38} & \num{105.05} & \num{2.48} & \num{1.75} & \num{3.70} & \num{117.43} \\
        Diffusion[0.5, 7] & \num{2.23} & \num{1.62} & \num{3.38} & \num{107.76} & \num{2.65} & \num{1.85} & \num{3.81} & \num{126.24} \\
        \bottomrule
    \end{tabular}
    \caption{Wind X 10M forecast metrics across different lead times.}
    \label{tab:wind_x_10m_metrics}
\end{table}
\begin{table}[h]
    \centering
    \setlength{\tabcolsep}{3pt} 
    \scriptsize 
    \begin{tabular}{l S S S S | S S S S }
        \toprule
        & \multicolumn{4}{c|}{\textbf{Lead Time: 60h}} & \multicolumn{4}{c}{\textbf{Lead Time: 120h}} \\
        \cmidrule(lr){2-5} \cmidrule(lr){6-9}
        & \multicolumn{1}{c}{Energy} & \multicolumn{1}{c}{CRPS} & \multicolumn{1}{c}{RMSE} & \multicolumn{1}{c|}{Spread} & \multicolumn{1}{c}{Energy} & \multicolumn{1}{c}{CRPS} & \multicolumn{1}{c}{RMSE} & \multicolumn{1}{c}{Spread} \\
        \midrule
        Deterministic & {-} & {-} & \num{2.33} & {-} & {-} & {-} & \num{4.06} & {-} \\
        Diffusion[0.5, 1] & \textbf{\num{1.19}} & \textbf{\num{0.92}} & \textbf{\num{2.32}} & \textbf{\num{67.35}} & \textbf{\num{2.04}} & \textbf{\num{1.61}} & \textbf{\num{3.68}} & \textbf{\num{107.12}} \\
        Diffusion[0.5, 3] & \num{1.65} & \num{1.19} & \num{2.91} & \num{85.30} & \num{2.39} & \num{1.76} & \num{3.87} & \num{118.32} \\
        Diffusion[0.5, 7] & \num{1.95} & \num{1.44} & \num{3.26} & \num{96.54} & \num{2.63} & \num{1.95} & \num{4.05} & \num{126.80} \\
        \midrule
        & \multicolumn{4}{c|}{\textbf{Lead Time: 180h}} & \multicolumn{4}{c}{\textbf{Lead Time: 234h}} \\
        \cmidrule(lr){2-5} \cmidrule(lr){6-9}
        & \multicolumn{1}{c}{Energy} & \multicolumn{1}{c}{CRPS} & \multicolumn{1}{c}{RMSE} & \multicolumn{1}{c|}{Spread} & \multicolumn{1}{c}{Energy} & \multicolumn{1}{c}{CRPS} & \multicolumn{1}{c}{RMSE} & \multicolumn{1}{c}{Spread} \\
        \midrule
        Deterministic & {-} & {-} & \num{4.27} & {-} & {-} & {-} & \num{4.58} & {-} \\
        Diffusion[0.5, 1] & \textbf{\num{2.46}} & \textbf{\num{1.74}} & \num{3.96} & \textbf{\num{124.53}} & \textbf{\num{2.41}} & \textbf{\num{1.63}} & \textbf{\num{4.07}} & \textbf{\num{125.59}} \\
        Diffusion[0.5, 3] & \num{2.68} & \num{1.80} & \num{3.99} & \num{131.01} & \num{2.74} & \num{1.83} & \num{4.17} & \num{132.56} \\
        Diffusion[0.5, 7] & \num{2.67} & \num{1.88} & \textbf{\num{3.94}} & \num{130.35} & \num{2.73} & \num{1.87} & \num{4.12} & \num{133.07} \\
        \bottomrule
    \end{tabular}
    \caption{Wind Y 10M forecast metrics across different lead times.}
    \label{tab:wind_y_10m_metrics}
\end{table}

\FloatBarrier
\section{Algorithms}\label{appendix:algos}

\begin{algorithm}[H]\label{alg:normalization}
\caption{Normalization Pipeline}
\begin{algorithmic}[1]
    \STATE \textbf{Input:} Data tensor $x_t$ and model $M_\theta$
    \STATE \textbf{Output:} Predicted tensor $y$
    \FOR{each variable $v$}
        \STATE $\mu_v = \frac{1}{h \times w}\sum_{i=1}^{h}\sum_{j=1}^{w} x_{t,v,i,j}$ 
        \STATE $\sigma_v = \sqrt{\frac{1}{h \times w}\sum_{i=1}^{h}\sum_{j=1}^{w} (x_{t,v,i,j} - \mu_v)^2}$ 
        \STATE $\hat{x}_{t,v} = \frac{x_{t,v} - \mu_v}{\sigma_v + \epsilon}$ 
    \ENDFOR
    \STATE $\hat{y} = M_\theta(\hat{x}_t)$ 
    \FOR{each variable $v$}
        \STATE $y= \hat{y} \times \sigma_v + \mu_v$ 
    \ENDFOR
    \STATE \textbf{Return:} $y$
\end{algorithmic}
\end{algorithm}

\begin{algorithm}[H]\label{alg:d_perturb}
\caption{Diffusion-based Perturbation Model: Inference}
\begin{algorithmic}[1]
    \STATE \textbf{Input:} Initial state $x_0$, Perturbation model $\epsilon_\phi$, Deterministic model $G_\theta$, number of samples $N$, number of trajectory steps $T$, guidance scale $\omega$
    \STATE \textbf{Output:} Trajectories $\{x_t^{(i)}\}_{t=1}^T, \, i = 1, \dots, N$
    \STATE Repeat input: $\{x_0^{(i)} = x_0\}_{i=1}^N$
    \FOR{$t = 1$ to $T$}
        \FOR{$i = 1$ to $N$}
            \STATE $z_S^{(i)} \sim \mathcal{N}(0, I)$
            \STATE Run DPM++ solver to sample perturbed state $\tilde{x}_t^{(i)}$ from $z_S^{(i)}$ using CFG:
            \FOR{reverse diffusion step $s = S, S{-}1, \dots, 1$}
                \STATE  $\hat{\epsilon}_u = \epsilon_\phi(z_s^{(i)}, \emptyset)$
                \STATE  $\hat{\epsilon}_c = \epsilon_\phi(z_s^{(i)}, x_{t-1}^{(i)})$
                \STATE $\hat{\epsilon} = (1 + \omega)\hat{\epsilon}_c - \omega \hat{\epsilon}_u$
                \STATE Update $z_{s-1}^{(i)}$ using DPM++ update rule with $\hat{\epsilon}$
            \ENDFOR
            \STATE $\tilde{x}_t^{(i)} \gets z_0^{(i)}$
            \STATE $x_{t+1}^{(i)} \gets G_\theta(\tilde{x}_t^{(i)})$
        \ENDFOR
    \ENDFOR
    \STATE \textbf{Return:} $\{x_t^{(i)}\}_{t=1}^T, \, i = 1, \dots, N$
\end{algorithmic}
\end{algorithm}

\begin{algorithm}[H]\label{alg:d_training}
\caption{Diffusion-based Perturbation Model: Training over Dataset}
\begin{algorithmic}[1]
    \STATE \textbf{Input:} Dataset $\mathcal{D} = \{x_0^{(i)}\}_{i=1}^N$, Diffusion model $\epsilon_\phi$, Deterministic model $G_\theta$
    \STATE \textbf{Output:} Trained perturbation model $\epsilon_\phi$
    \FOR{each sample $x_0 \in \mathcal{D}$}
        \STATE $s \sim \text{Bernoulli}(0.5)$
        \IF{$s = 1$}
            \STATE $x_0 \gets G_\theta(x_0)$ \COMMENT{Advance sample through time}
        \ENDIF
        \STATE $t \sim \mathcal{U}\{1, \dots, T\}$
        \STATE  $\epsilon \sim \mathcal{N}(0, I)$
        \STATE $x_t = \sqrt{\bar{\alpha}_t} x_0 + \sqrt{1 - \bar{\alpha}_t} \epsilon$
        \STATE $g \sim \text{Bernoulli}(\lambda)$
        \IF{$g = 1$}
            \STATE $\hat{\epsilon} = \epsilon_\phi(x_t, c)$ \COMMENT{Use conditional input $c$}
        \ELSE
            \STATE $\hat{\epsilon} = \epsilon_\phi(x_t, \emptyset)$ \COMMENT{Use unconditional input}
        \ENDIF
        \STATE $\mathcal{L} = \|\hat{\epsilon} - \epsilon\|^2$
        \STATE $g \gets \nabla_\phi \mathcal{L}$
        \STATE $g \gets \frac{g}{\max(1, \|g\|_2 / \tau)}$ \COMMENT{Gradient norming with threshold $\tau$}
        \STATE $\phi \gets \phi - \eta \cdot g$
    \ENDFOR
\end{algorithmic}
\end{algorithm}

\end{document}